\documentclass[10pt,letterpaper]{article}
\usepackage[top=0.85in,left=2.75in,footskip=0.75in]{geometry}

\usepackage{amsmath,amssymb}

\usepackage{changepage}

\usepackage[utf8x]{inputenc}

\usepackage{textcomp,marvosym}

\usepackage{subfigure}
\usepackage{lineno,hyperref}

\usepackage{amssymb}
\usepackage{amsmath}
\usepackage{siunitx}
\usepackage{tabularx}
\usepackage{algorithmic}

\usepackage{MnSymbol,wasysym}
\usepackage{algorithm2e} 
\usepackage{amsthm}
\usepackage{graphicx}
\usepackage{lscape}
\usepackage{verbatim}
\usepackage{color,soul}
\usepackage{xcolor}
\usepackage{subfigure}
\usepackage{tabularx,ragged2e,booktabs,caption,array,multirow,multicol}
\usepackage{csquotes}
\usepackage{mathtools}
\usepackage{lineno}
\usepackage{amsthm}  
\usepackage{listings}
\usepackage{amssymb}
\usepackage{latexsym}
\usepackage{epsfig}
\usepackage{float}
\usepackage{lscape}
\usepackage{xspace} 

\usepackage{tikzsymbols}
\usepackage{subfigure} 
\usepackage{rotating}
\usepackage{cite}

\usepackage{epstopdf}
\AppendGraphicsExtensions{.tiff}

\usepackage{lineno}

\usepackage{microtype}
\DisableLigatures[f]{encoding = *, family = * }

\usepackage{array}

\newcolumntype{+}{!{\vrule width 2pt}}

\newlength\savedwidth

\raggedright
\usepackage[aboveskip=1pt,labelfont=bf,labelsep=period,justification=raggedright,singlelinecheck=off]{caption}

\makeatletter
\renewcommand{\@biblabel}[1]{\quad#1.}
\makeatother

\usepackage{lastpage,fancyhdr,graphicx}
\usepackage{epstopdf}
\fancyheadoffset[L]{2.25in}
\fancyfootoffset[L]{2.25in}
\begin{document}
\vspace*{0.2in}

\begin{flushleft}
{\Large
\textbf\newline{COVID-19 sentiment analysis via deep learning during the rise  of novel cases}  
}
\newline 
\\ 
Rohitash Chandra \textsuperscript{1 \ddag *}, Aswin Krishna \textsuperscript{2 \ddag } 
\\
\bigskip 

 1. School of Mathematics and Statistics, University of New South Wales, Sydney, Australia
\\
2. Department of Chemical Engineering, Indian Institute of Technology Guwahati, Assam, India
\\
\bigskip

%
%




\ddag These authors contributed equally to this work.

* Corresponding author\\
E-mail: rohitash.chandra@unsw.edu.au (RC)

\end{flushleft}
\section*{Abstract}
  
Social scientists and psychologists  take  interest  in understanding   how people express emotions and  sentiments  when dealing with catastrophic events  such as natural disasters, political unrest,  and terrorism. The COVID-19 pandemic is a  catastrophic event that has raised a number of psychological issues such as depression given abrupt social changes and lack of employment. Advancements of deep learning-based language models have been promising for sentiment analysis with data from social networks such as Twitter. Given the situation with  COVID-19 pandemic, different countries had different peaks where rise and fall of new cases affected lock-downs which directly affected the economy and employment. During the rise of COVID-19 cases with stricter lock-downs, people have been expressing their sentiments    in social media. This   can provide a deep understanding of human   psychology  during catastrophic events.     In this paper,  we present a framework that employs  deep learning-based language models via long short-term memory (LSTM) recurrent neural networks for sentiment analysis during  the rise  of  novel COVID-19 cases in India. The framework features LSTM language model with a global vector   embedding and   state-of-art BERT language model. We  review the   sentiments expressed  for selective months in 2020 which covers the first major peak of novel cases in India. Our  framework  utilises multi-label sentiment classification  where more than one sentiment can be expressed at once. Our  results indicate that the majority of the tweets have been  positive with high levels of optimism during the rise of the novel COVID-19 cases   and the number of tweets significantly lowered towards the peak. We find   that the optimistic, annoyed and joking tweets   mostly dominate  the monthly tweets with much lower portion  of negative sentiments.  The predictions generally  indicate that although the majority have been optimistic, a significant  group of population has been annoyed towards the way the pandemic was handled by the authorities.



 
\section{Introduction}
\label{S:1}

There has been unprecedented growth of information   via social media which has been of interest to social scientists and psychologists in a better understanding of the human condition, psychology  \cite{golbeck2011predicting,quercia2011our,bittermann2021mining,lin2015building} and mental health \cite{coppersmith2014quantifying}. Social media platforms such as  Twitter  has been used as a medium for    data acquisition for research in  psychology and behavioural  science  \cite{murphy2017hands}. It  has also been used as a tool for predicting personality type \cite{golbeck2011predicting,quercia2011our}, and understanding trends and backgrounds of users online \cite{zhou2019comparative,wang2016twitter}. There has also been interest   regarding how people express sentiments when dealing with catastrophic events  such as natural disasters,  extreme political viewpoints \cite{alizadeh2019psychology},  and terrorism   \cite{garg2017sentiment}. For instance, in the case of a terror attack in Kenya \cite{garg2017sentiment},   Twitter became a crucial channel of communication between the government, emergency response team and the public.  

Sentiment analysis  \textcolor{black}{involves the use of} natural language processing (NLP) \cite{manning1999foundations} methods  to systematically   study affective states and emotion understanding of individuals or social groups \cite{liu2012survey,medhat2014sentiment,hussein2018survey}. \textcolor{black}{We note that deep learning, which is a machine learning method, has been prominently used for NLP tasks.}
Apart from research in psychology, sentiment analysis has a number of applications such as  understanding customer behaviour \cite{ordenes2014analyzing},  clinical medicine \cite{greaves2013use}, building better prediction model for trading stocks \cite{mittal2012stock}, and elections such as the US Presidential campaign in 2012 \cite{wang2012system}. \textcolor{black}{Recently, there has been a trend of using deep learning-based language models   \cite{zhang2018deep}  with training data from   
Twitter  for  sentiment analysis} \cite{kouloumpis2011twitter,giachanou2016like}. One of the earliest works began using NLP methods such as n-grams  with hash tags for building training data and   machine learning methods such as Adaboost  for sentiment classification \cite{kouloumpis2011twitter}. Deep learning methods such as convolutional neural networks have been used for sentiment analysis on Twitter \cite{severyn2015twitter}. 
 

 The \textit{coronavirus disease 2019} (COVID-19) \cite{Gorbalenya2020species,monteil2020inhibition,world2020coronavirus,cucinotta2020declares} global  pandemic has been a catastrophic event with major impact on the world's economy which created   rise in  unemployment,   psychological issues, and  depression. The abrupt social, economic and travel changes has motivated research from various fields \cite{siche2020impact,richards2020impact,tiwari2021delhi}, where computational   modelling with machine learning has been prominent \cite{shinde2020forecasting}; however, it  had a number of  challenges due to the testing and reporting of cases \cite{rahimi2021review}. Deep learning models  have played a significant role in forecasting COVID-19 infection treads   for various parts of the world \cite{zeroual2020deep,ChandraCOVID2021,TiwariAir2021}.  
 
 During rise of COVID-19 cases, and stricter lock downs, people have been expressing different sentiments in social media such as Twitter. Social media has played a significant role during COVID-19 which has driven researchers for analysis with NLP and machine learning methods. 
A study that used  sentiment analysis   via deep learning  reported that  World Health Organisation (WHO)   tweets have been unsuccessful in providing public guidance \cite{chakraborty2020sentiment}. There has been a study on sentiment analysis to study the effect of  nationwide lockdown due to COVID-19 outbreak in India where it was  found that people took the fight against COVID19 positively and majority were in agreement with the government for  the initial nation-wide lockdown  \cite{barkur2020sentiment}.  Social media  posts and tweets brings another level of understanding when combined with sentiment analysis. An example of topic modelling examined  tweets during the COVID-19 pandemic  identified    themes such as 'origin of the virus',    and 'the economy' \cite{abd2020top}. Topic modelling has been used with Twitter based sentiment analysis during early stages of COVID-19 and sentiments such as fear was dominant \cite{xue2020public}. In region specific studies, tweets  from the United States  was used to determine the   network of dominant topics and   sentiments  \cite{hung2020social}. Further work was done  in case of  China via \textit{bi-directional
 encoder representations from transformers} (BERT) language  model for trend, topic, and   sentiment analysis \cite{wang2020covid}.  Further region specific studies include community sentiment analysis in Australia \cite{zhou2020examination} and    sentiment analysis in Nepal \cite{pokharel2020twitter},    where  majority positive sentiments were found with elements of fear.  In Spain, sentiment analysis  reviewed  how  digital platforms created an impact during COVID-19 \cite{de2020sentiment}. A study of 
   cross-language sentiment analysis of European Twitter messages during the first few months of the COVID-19 pandemic  found that the  lockdown announcements correlate with a deterioration of moods, which recovers within a short time span \cite{kruspe2020cross}.

 
   The sentiment analysis examples from Europe and India show how social media can play a powerful role in understanding  the psychology  \textcolor{black}{and the human condition} Recent advancements of deep learning models as a tool for building robust language models have provided further  motivation in understanding the temporal nature of the sentiments during and after the first peak of COVID-19 in  India. Different countries had different peaks where rise and fall of new cases implemented lock-downs which directly affected the economy and employment. India is special in this way (until February 2021), where a single nation-wide peak was seen and only certain states had multiple peaks  (Delhi and Maharashtra)  \cite{ChandraCOVID2021}.

 In this paper,  we use deep learning based language models via long short-term memory (LSTM) recurrent neural networks for sentiment analysis  via  tweets with a focus of rise  of  novel cases in India. \textcolor{black}{We use LSTM and bidirectional  LSTM (BD-LSTM) model with global vector (GloVe) for word representation  for  building a language model. Moreover, we use the BERT model to  compare the results from LSTM and BD-LSTM models and then use the best model for COVID-19 sentiment analysis for the case of India.}   We use three datasets, which include India, along with the state of Maharashtra (includes Mumbai) and Delhi. We  compare the monthly sentiments expressed  covering the major peak of new cases in 2020. We present  a framework that focuses on multi-label sentiment classification, where more than one sentiment can be expressed at once. We use Senwave COVID-19 sentiment dataset \cite{yang2020senwave} which features 10,000 tweets collected worldwide and hand-labelled by 50 experts  for training  LSTM models.
 
 \textcolor{black}{We highlight that there is no study that uses language models for sentiment analysis during the rise of novel COVID-19 cases. Our framework compares the different types of sentiments expressed across the different months in relation to the rise of the number of cases, which impacted the economy and had different levels of lock downs. This had an effect on the psychology of the population given  stress and fear. Hence,  the study is a way to quantify and validate emotional and psychological conditions  given uncertainty about the pandemic. The major contribution of the paper is in using state-of-art sentiment analysis methods for understanding the public behaviour in terms of psychological well being   in relation to the rise  of novel COVID-19 infections. This study presents a novel framework that makes use of  information from social media for understanding public behavior during  a major disruptive event of the century.}

The rest of the paper is organised as follows: Section 2 presents a background   of related work, and  Section 3 presents the proposed methodology  with data analysis. Section 4 presents experiments and results. Section 5 provides a discussion and Section 6 concludes the paper with discussion of future work. 

\section{Related Work}

Word embedding is the process of feature extraction from text for NLP tasks such as sentiment analysis \cite{li2018word,kutuzov2018diachronic,ruder2019survey}. Word embedding can be obtained using methods where words or phrases from the vocabulary are mapped to vectors of real numbers. The process generally involves a mathematical embedding from a large corpus  with many dimensions per word to a vector space with a   lower dimension  that is useful for machine learning or deep learning models for text classification tasks \cite{li2018word}. Basic word embedding methods such as \textit{bag of words} \cite{zhang2010understanding} and \textit{ term frequency inverse document
frequency } \cite{ramos2003using} do not have context awareness and semantic information  in embedding.  This is also a problem for skip-grams \cite{goodman2001bit} that use n-grams  (such as bigrams and  tri-grams) to develop word embedding, and in  addition allow  adjacent sequences of words  tokens to be “skipped” \cite{guthrie2006closer}.  
 
Over the last decade, there has been  phenomenal progress in the area of world embedding and language models. Mikolov et al.  \cite{mikolov2013distributed}  proposed \textit{word2vec} embedding which uses a feedforward  neural network model to learn word associations  from a text dataset which can detect synonymous words or suggest additional words given  a partial sentence. It uses  continuous bag-of-words (CBOW)   or continuous skip-gram model   architectures to produce a distributed representation of words.  The  method is used to create a large  vector which represent each unique word  in the corpus where semantic information and relation  between the words are preserved.  It has been shown that for  two sentences  that do not have much words in common, their semantic similarity can be captured using word2vec \cite{mikolov2013distributed}. The limitation of word2vec is that it does not well represent the context of a word.  \textcolor{black}{Pennington et al. \cite{pennington2014glove} proposed global vectors   (GloVe) for word embedding  as an unsupervised learning algorithm} for obtaining vector representations for words by mapping words into a meaningful space where the distance between words is related to semantic similarity . GloVe uses matrix factorization to constructs a large matrix of   co-occurrence information to obtain representation that showcase  linear substructures of the word vector space. The  embedding  feature vectors with top list words that match with certain distance measures. GloVe \textcolor{black}{can be used to find relations} between words such as  synonyms, company-product relations.  \textcolor{black}{Due to the awareness in ethics in machine learning, there has been a major focus on ethical issues in NLP. } A recent study showed that GloVe can have gender biased information; hence, a gender neutral GloVe method has been proposed \cite{zhao2018learning}.

There are some studies that review the effectiveness of word embedding methods. Ghannay et al. \cite{ghannay2016evaluation} \textcolor{black}{provided an evaluation} of word embedding methods  such as GloVe  \cite{pennington2014glove}, skip-gram, and continuous space language models (CSLM)  \cite{schwenk2007continuous}. The authors reported that skip-gram and GloVe   outperformed CSLM in all the language tasks. Wang et al. \cite{wang2018comparison} evaluated word embedding methods such as GloVe for  applications of  biomedical text analysis where it was found that  word embedding  trained from clinical notes and literature better captured word semantics.

\section{Methodology}

 \subsection{LSTM and BERT language models}
 
     Recurrent neural networks (RNN)  in general feature  a  context memory layer to incorporate previous state and current  inputs for propagating information to the next state, and eventually the output layer for decision making. Canonical RNNs (simple RNNs) feature  several different architectures, besides the Elman RNN \cite{elman_Zipser1988,Elman_1990}   for modelling temporal sequences  and dynamical systems \cite{Omlin_etal1996,Omlin_Giles1992,Chandra_Omlin2007}.   One of the major challenges   in training simple RNNs is due to the architectures properties of unfolding in time for long-term dependency problems. Backpropagation through time (BPTT), which is an extension of the backpropagation algorithm, has been prominent for training simple RNNs \cite{Werbos_1990}.  Due to problem of learning long-term dependencies given  vanishing and exploding gradients  with simple RNNs \cite{Hochreiter_1998}, long short-term memory (LSTM) recurrent neural networks have been developed \cite{hochreiter1997long}. LSTM networks have  better capabilities  for learning long-term dependencies in temporal data  using memory cells and gates.  

In the last decade, with the deep learning revolution, several LSTM architectures have been developed. \textcolor{black}{ Bidirectional LSTM models \cite{graves2005} process information in two directions, rather than making use of only previous context state for determining the next  states which are based on bidirectional RNNs  \cite{schuster1997}}. In this way,
  two independent LSTM models allow  both backward and forward information about the sequence at every time step.  This enables  better access to \textcolor{black}{long range state information  which have been useful for word embedding   \cite{graves2005} and    several other sequence processing problems \cite{graves2005,Fan2014TTSSW,graves2013hybrid}.}

  \textcolor{black}{A transformer is an extended LSTM model that adopts the mechanism of attention which mimics cognitive attention to enhance important parts of the data while fading the rest \cite{vaswani2017attention}. Transformers also use an encoder-decoder architecture and have mostly been used for  NLP tasks  such as translation and text summarising \cite{vaswani2017attention,wolf2020transformers}. In comparison to conventional  RNNs, transformers do not require data to be processed in a sequential  order since the attention operation provides context for any position in the input sequence.  BERT is a pre-trained  transformer-based model for NLP tasks which has   been used  to better understand user behaviour in Google search engine \cite{devlin2018bert}. BERT can be used for a number of other  NLP applications such as clinical data processing  \cite{su2020application}. The original BERT \cite{devlin2018bert} has two models:  1.)  BERT-base features  12 encoders with 12 bidirectional self-attention heads, and 2.) BERT-large features 24 encoders with 16 bidirectional self-attention heads. These  are pre-trained from unlabeled data extracted from a corpus with 800 million words and English Wikipedia with 2,500 million words, respectively. Word2vec and GloVe are content-free models that generate a single word embedding representation for each word, whereas BERT takes into account the context for each occurrence of a given word which makes BERT one of the best language models. Hence, BERT is suitable for our sentiment analysis tasks using tweets during rise of COVID-19 cases in India. }

  \subsection{Twitter-based Sentiment Analysis Framework for COVID-19}

\begin{figure*}[htbp!]
\centering
\includegraphics[width=14cm]{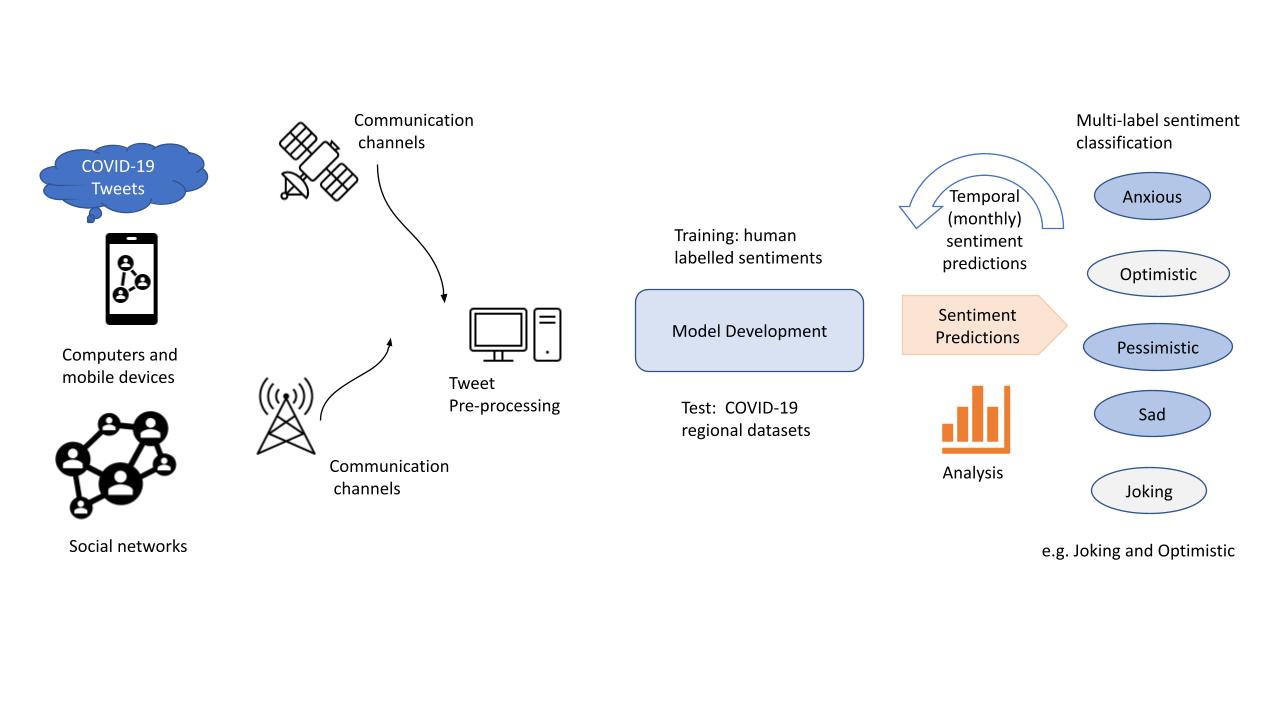}
\caption{\textcolor{black}{Framework: Twitter-based sentiment analysis   for COVID-19 with  model development  via LSTM, BD-LSTM and BERT. After training, the best model is chosen for prediction of COVID-19 tweets from India.} Note that the framework features multi-label classification which has the ability to provide more than two outcomes at once; i.e. a tweet can be optimistic and joking at the same time. }
\label{fig:framework}
\end{figure*}

 Our framework for sentiment analysis with deep learning follows a series of steps with major components that involve: 1.) tweet extraction; 2.) tweet pre-processing; 3.)  model development and training using LSTM, BD-LSTM and BERT; and 4.) prediction using selected COVID-19 data. Figure \ref{fig:framework} shows the framework diagram where we highlight  the major components. We note that the framework features multi-label classification which has multiple outcomes at once, i.e the tweet can be optimistic and joking at the same time. 
 
 The first step involves processing on COVID-19 dataset from a selected region. \textcolor{black}{We choose COVID-19 tweets from while of  India along with   two states which have been amongst the highest states of COVID-19 cases in 2020 (Maharashtra and Delhi). } \textcolor{black}{We process Twitter data obtained for India and related state as follows. We note that special phrases, emotion symbols (emoji's) and abbreviations are present in tweets, since language in social media has been rapidly evolving. Hence, we have to transform them for building the language model. Table \ref{tab:process} shows other words and emotions that are transformed.} We note that since this is a case of India, where English is used in combination with some key indigenous Indian languages (such as Hindi). Hence, we  transform certain words, emotions and character symbols expressed in other languages as well. 
 Note that we use hand-labelled sentiment dataset (Senwave COVID-19 dataset) \cite{yang2020senwave} which features  11 different sentiments labelled by a group of  50 experts for 10,000 tweets worldwide during COVID-19 pandemic in 2020. In this dataset, "official report" is classified as a sentiment although it is a topic.

\begin{center}
\begin{table}[htbp!]
    \small
    \begin{tabular}{l l } 
    \hline
    Original Phrase/Emoji & Transformed Word  \\ 
    \hline
    
    omg & oh my god  \\ 
    
    btw & by the way   \\
    
    socialdistancing & social distancing  \\
     \smiley{} & smiling face     \\
     \frownie{} & sad face    \\
     \Bed & bed  \\
     \Fire &  fire   \\ 
    \Winkey & wink  \\
    \Laughey & laugh  \\
    \hline
    \hline
    \end{tabular}
    \caption{Examples of word transformation that features changing symbols and emoji's to words that can have semantic representation.   }
    \label{tab:process}
\end{table}
\end{center}

  The next step involves converting each word into its corresponding GLoVe embedding vector, where each word is converted into a vector (300 dimensions in our case). We select GLoVe embedding since it has shown good results for language models with sentiment analysis in the literature \cite{pennington2014glove}. The aforementioned vector from each word thus obtained is passed on to the respective LSTM models for training once the model architecture has been defined as shown in Figure \ref{fig:framework}.   We  first evaluate our  trained models (LSTM, BD-LSTM, and BERT) and then  use it for COVID-19 sentiment analysis using test data (India, Maharashtra, and Delhi). 
 As shown in Figure \ref{fig:framework}, the trained model is used to classify 11 sentiments such as  ``anxious'',  ``sad'', and   ``joking''.

 It is essential to have the right metric to evaluate the model for the given application.  The outcome of a  misclassification in a multi-label classification  is no longer a  correct or an incorrect  instance as opposed to binary classification or multi-class classification \cite{dembczynski2012label}. In multi-label classification, a prediction that features  a subset of the actual classes are  considered better than a prediction that contains none of the actual classes; hence, this needs to be captured by the metric that captures the loss or gives a score.   As given in the literature \cite{dembczynski2012label}, multi-label classification evaluation is typically based on 1.) binary cross-entropy (BCE) loss\cite{zhang2018generalized} which  is a softmax activation combined with a cross-entropy loss; 2.) Hamming loss\cite{dembczynski2012label} which   calculates loss generated in the bit string of class labels  using \textit{exclusive or} (XOR) operation  between the actual and predicted labels and then computes the  average over  the instances of the dataset. The Jaccard coefficient score\cite{hamers1989similarity}     provides a  measure of the overlap between actual and predicted labels  with their attributes capturing  similarity and diversity. Furthermore, label ranking average precision (LRAP) score \cite{furnkranz2008multilabel} finds the percentage  of the higher-ranked labels that  resemble  true labels  for each of the given samples. F1-score  conveys the balance between the precision and the recall. F1-score has been prominent for understanding class imbalance problems \cite{jeni2013facing}. \textcolor{black}{In application to multi-label classification, two types of F1-scores are predominantly used. F1-macro is computed using the F1-score per class/label and then averaging them, whereas F1-micro computes the F1-score for the entire data considering the total true positives, false negatives and false positives} \cite{lewis1996training}. We   use a combination of these scores to evaluate our model results for the first phase of the framework where the respective  models are trained.

\subsection{Model Training}

We present an experimental study that  compares multi-label classification using \textcolor{black}{LSTM, BD-LSTM, and BERT models} as shown in the framework  (Figure \ref{fig:framework}). We use 90\% of the dataset for training and 10\% for testing.  We use the respective models and train them using the Senwave COVID-19 dataset which is publicly available, but restricted and needs permission from the authors for usage  \cite{yang2020senwave}. Therefore, we cannot provide the processed dataset, but rather provide the trained  models via Github repository (\url{https://github.com/sydney-machine-learning/COVID19_sentinentanalysis}). The Senwave dataset features 10,000 Tweets collected from March to May 2020. 

\textcolor{black}{In the LSTM model, we determine the model hyper-parameters based on how the model performs in trial experiments. The  GloVe embedding uses a  word vector of size  300 size to provide data representation which is a standard representation in the literature  \cite{pennington2014glove}. We use a dropout  regularisation probability of 0.65 for LSTM and BD-LSTM models which feature: 300 input units,  two layers with 128  and 64 hidden units,  and finally an output layer with 11  units for classifying the sentiments. In the case of  BERT,  we use the default hyper-parameters for the BERT-base (uncased) model and only tune the learning rate. The only major addition  is a dropout layer at the end of the BERT architecture followed by a linear activation layer with 11 outputs corresponding to the 11 sentiments.}
 
   Table \ref{tab:train} shows the model training results  (mean)  for  10 experiments with different initial weights and biases for the respective models with different performance metrics. \textcolor{black}{Generally, we find that BERT is better than LSTM and BD-LSTM models across the different performance metrics shown in Table \ref{tab:train}.  Hence, we proceed with the BERT  model for evaluating the test dataset  with a visualization of the predictions.}

        \begin{center}
        \begin{table}[htbp!]
            \small
            \begin{tabular}{c c c c} 
            \hline
            Metric & BD-LSTM & LSTM & \textcolor{black}{BERT} \\ 
            \hline
            \hline
            BCE Loss & 0.281 & 0.255 & 0.372\\ 
          
            Hamming Loss & 0.163 & 0.157 & 0.142\\
             
            Jaccard Score & 0.417 & 0.418 & 0.510\\
            
            LRAP Score & 0.503 & 0.511 & 0.766\\
           
            F1 Macro & 0.434 & 0.430 & 0.530\\ 
          
            F1 Micro & 0.495 & 0.493 & 0.587\\ 
            \hline
              \hline
            \end{tabular}
            \caption{Training performance for \textcolor{black}{BD-LSTM, LSTM and  BERT model} using Senwave COVID-19 training dataset. Note that except for the BCE and Hamming loss, higher scores shows better performance.}
            \label{tab:train}
        \end{table}
        \end{center}
 
 \subsection{Application to COVID-19 pandemic in India}

  India  COVID-19 pandemic management had major challenges given large population with a large number of  densely populated cities \cite{lancet2020india}.   India reported first COVID-19 case on 30th January 2020 and entered a lock-down afterwards which was gradually changed.
India  (22-nd March,  2021) had more than 11.6 million  confirmed cases with more than 160 thousand  deaths which made India the third highest  confirmed cases in the world after the United States and Brazil. India was the   8th in world with more than 300 thousand active cases (prior to the second wave). India had a single major peak around middle of September in 2020 with close to 100 thousand cases daily, which gradually decreased to around 11 thousand cases daily end of January 2021, and since then it was gradually raising. In March 2021, the cases began rising faster and by 22nd March, India had 47 thousand daily new cases which was moving  towards a second peak \cite{indiaworld}. 

In the first six months, the states of Maharashtra (population of about 124 million), Delhi (population of 19 million), and Tamil Nadu (population of  about 8 million) led COVID-19 infections \cite{indiapop}.  \textcolor{black}{Note that the city of Mumbai is in the state of Maharashtra, which in terms of the   population can be compared to some } of the  highly  populated countries.  In the later half of the year, Delhi cases reduced but still remained in leading 8 states \cite{ChandraCOVID2021}. In 2021, Maharashtra continued as state of highest infections and in March, it was featuring more than half of new cases on a weekly basis and Delhi   contained the situation with less than a thousand daily cases. Hence, our focus is to study whole of India with two states of interest that includes Maharashtra  and  Delhi.

 We note that the proposed framework can be applied to any part of the world; however, we are choosing the case of India to show the effectiveness of the framework.  
 The final step is in applying the different datasets from COVID-19 in India which include, nation-wide COVID-19 sentiments, and two major states with COVID-19 cases.   The trend in the cases shows that both states had a major peak followed by minor ones, whereas India as a whole had a single major peak which was around mid-September 2020 with close to 97,000 novel cases per day during the peak as shown in the next section  (Figure 2).

\section{Results}


In this section, we provide results of the implementation of our proposed framework  using Twitter dataset for COVID-19 in India.
 
\subsection{COVID-19   Data Visualisation}

Our test dataset \cite{lamsal2020design} features COVID-19 India which features tweets associated with COVID-19 from March to September 2020. It consists of more than 150,000 tweets from  India. We generate two other datasets from this which feature the regions  of Maharashtra (state) and Delhi (union territory) which contains around 18,000 tweets each, respectively.  

\begin{figure}[htbp!]
\centering
\subfigure[India]{
\includegraphics[scale =0.25]{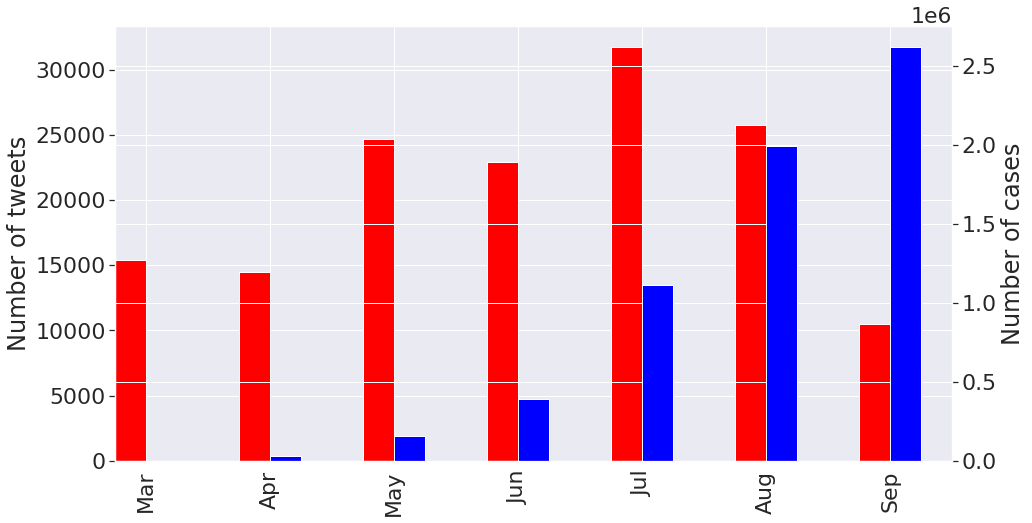}
 }
 \subfigure[Maharashtra]{
   \includegraphics[scale =0.25]{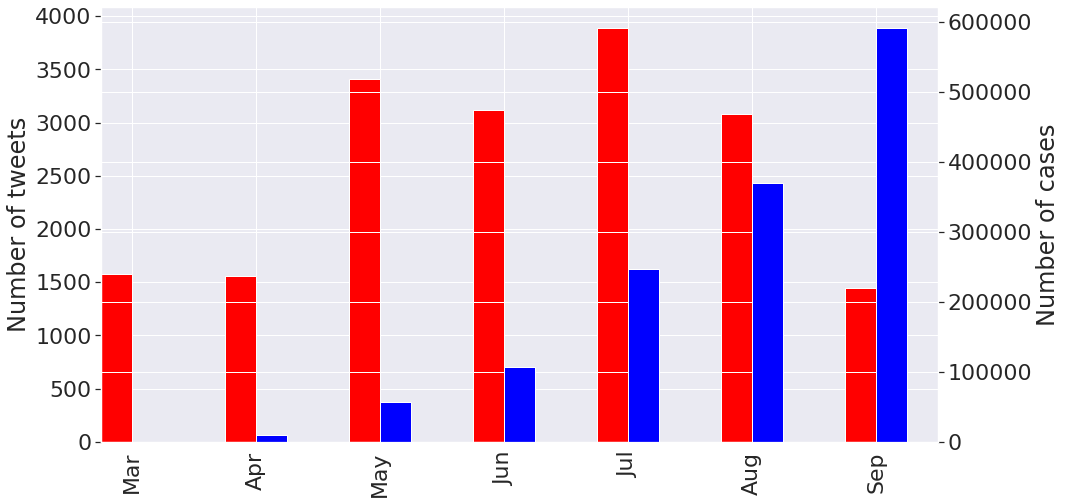}
 } 
  \subfigure[Delhi]{
   \includegraphics[scale =0.25]{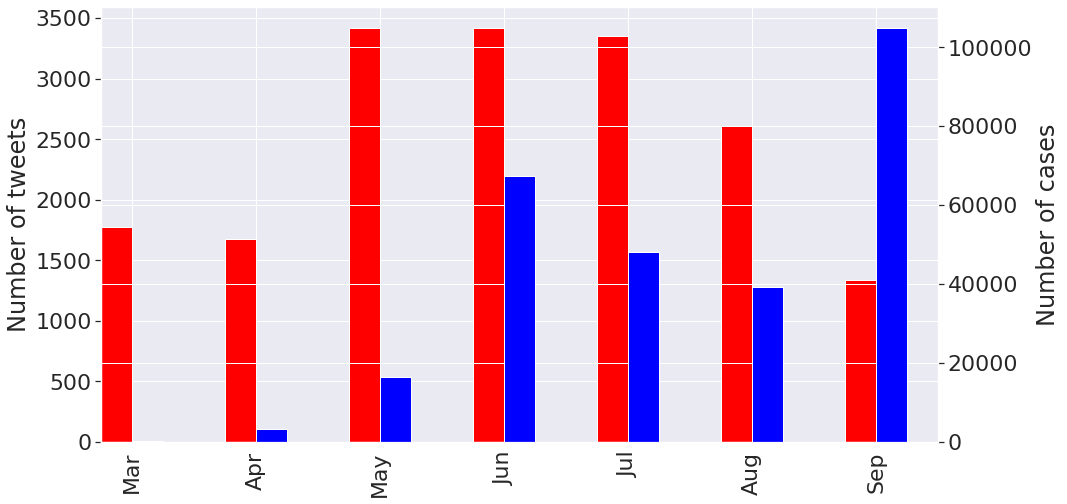}
 } 
 
\caption{Novel COVID-19 tweets and cases in India, Maharashtra, and Delhi. \textcolor{black}{The red bars indicate the number of tweets while the black bars indicate the  number of novel cases.} }

\label{fig:cases}
\end{figure}

We first provide visualisation of the dataset and report features of interest to language models such as bi-grams, and tri-grams and distribution of monthly tweets. Figure \ref{fig:cases} shows the distribution of number of tweets for selected months along with the number of cases for the respective datasets. 

We notice that the number of tweets in the case of India follows the same trend as the number of novel COVID-19 cases until July, afterwards the number of tweets decline  (Panel~a,   Figure~\ref{fig:cases}). There is a similar pattern for case of Maharashtra (Panel~b). The case of Delhi (Panel~c) is slightly different as the first peak was reached in July with increasing tweets that declined afterwards and did not keep up with the second peak of cases in September.  This indicates  that as more cases were observed in the early months, there was much concern which eased before the major  peak was reached and the number of tweets were drastically decreased. There could be elements of fear, depression and  anxiety as the tweets decreased drastically after July with increasing cases, which we analyse in the following section.  

\begin{figure}[htbp!]
\centering
\subfigure[Bi-grams]{
\includegraphics[scale =0.33]{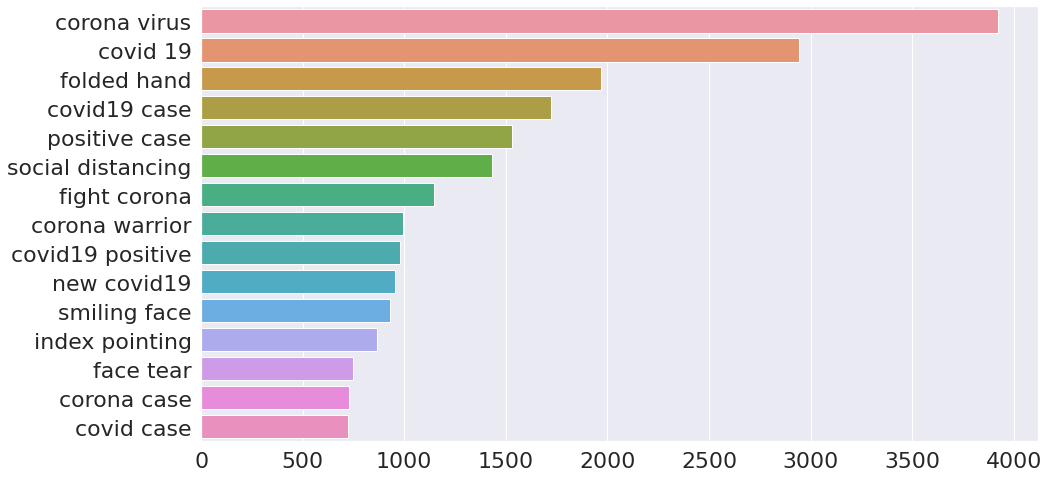}
 }
 \subfigure[Tri-grams]{
   \includegraphics[scale =0.33] {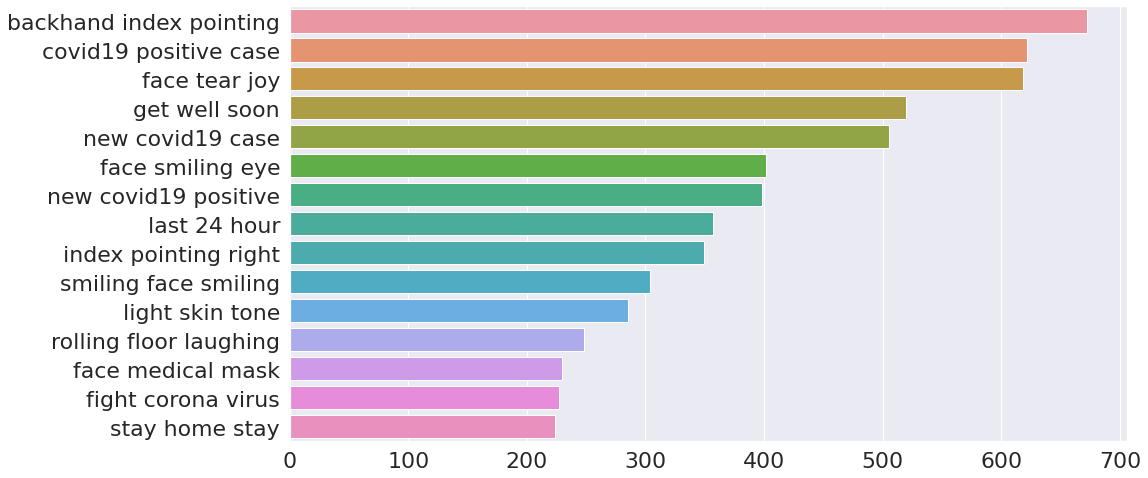}
 }  
 
\caption{Bi-grams and tri-grams for case of India. Note that a combination of emoji's such as ``backhand - index - pointing", ``smiling - face",   and words have been used. }

\label{fig:ngrams_india}
\end{figure}

Figure \ref{fig:ngrams_india} shows a visualisation of the number of bi-grams and tri-grams for the case of India. In case of bi-grams (Panel a), we find that the term ``corona - virus" is mostly used followed by ``covid - 19" which are essentially  the same. Next, it is interesting that the terms ``folded - hand" are mostly used followed by ``positive - case" and ``social - distancing". The term  ``folded - hand" shows Indian social and religious symbol that denotes keeping the faith, giving respect and also a sign of acknowledgement and appreciation. We note that the ``folded hand" is an emotion icon (emoji) used predominantly in social media, and hence been processed as a text during prepossessing (Table 1), which is actually is not a bi-gram from a semantic viewpoint.   In order to provide  better  understanding of the context of the tweets,    we give examples  in Table \ref{tab:bigrams}. In the case of tri-grams (Panel b), we find more information in tweets such as "backhand - index - pointing" which is an emoji; hence, we provide some of the tweets that fall in this category in Table \ref{tab:trigrams}.
 
\begin{figure}[htbp!]
\includegraphics[width=10cm]{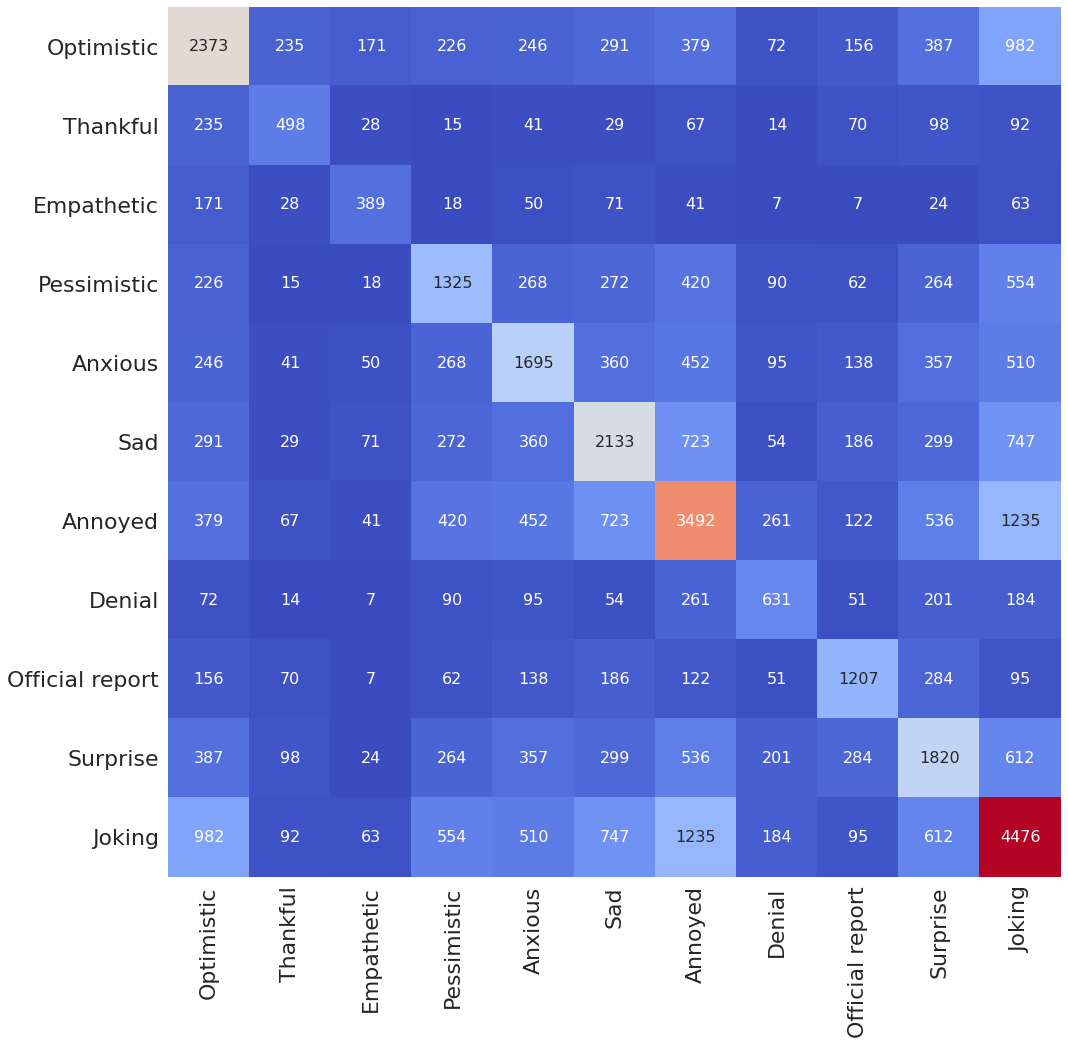}
\caption{Heatmap  showing number of occurrence  of a given sentiment in relation to the rest of the sentiments for 10,000 tweets from Senwave dataset \cite{yang2020senwave} used for training.}
\label{fig:heatoccur_senwave}
\end{figure}

   Figure  \ref{fig:heatoccur_senwave}  shows number of occurrence  of a given sentiment in relation to the rest of the sentiments  for 10,000  hand-labelled tweets  in Senwave dataset \cite{yang2020senwave} used for training.

    \begin{table*}[htbp!]
        \small
       \centering
        \begin{tabular}{l l l} 
        \hline
        Month & Tweet & Bi-gram (Emoji) \\ 
        \hline
             \hline
       March & ``releasing today at 6:00 pm on my youtube channel!   & ``folded hand" \\
       
       &   let's fight this together folded hands i need your support guys" &  \\
        \hline
       July & ``that's really shameful and heinous folded hands" & ``folded hand" \\
        \hline
       August & ``please applaud this corona economy warrior. folded hands kudos." & ``folded hand" \\
        \hline
       March & ``india this backhand index pointing down" & ``index pointing" \\
        \hline
       May & ``corona : to everyone backhand index pointing down" & ``index pointing" \\
          \hline
       July & ``backhand index pointing right. ........lockdown time.....  & ``index pointing" \\
         
    &   \#picture \#instagood \#instadaily \#insta" &  \\
       
        \hline
             \hline
        \end{tabular}
        \caption{Selected examples of processed tweets that are captured in most prominent bi-grams.}
        \label{tab:bigrams}
    \end{table*}

    \begin{table*}[htbp!]
        \small
       \centering
        \begin{tabular}{l l l} 
        \hline
        Month & Tweet &  Tri-gram (Emoji) \\
        \hline
         \hline
        May & ``google doc of resources for amphan and covid.    & ``backhand - index - pointing"  \\
         & backhand index pointing down.    &   \\
         &   retweet for greater reach." &   \\
        \hline
        August & "\#covid tips backhand index pointing down" &  ``backhand - index - pointing" \\
        \hline
        September & "backhand index pointing right registers   & ``backhand - index - pointing" \\
        
     &   highest single-day cases in the world - 95,529" & \\
        \hline
         \hline
        
        \end{tabular}
        \caption{Selected examples of processed tweets that are captured in most prominent tri-grams.}
        \label{tab:trigrams}
    \end{table*}

\subsection{Sentiment Prediction}

\begin{figure}[htbp!]
\centering
\subfigure[India]{
\includegraphics[scale =0.27]{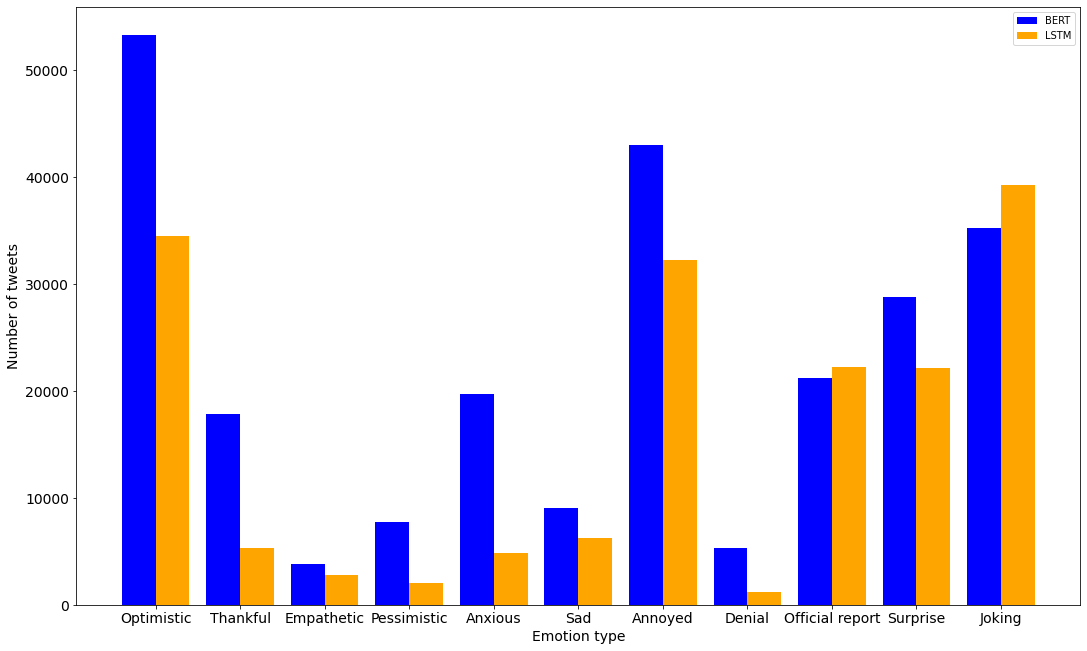}
 }
 \subfigure[Maharashtra]{
   \includegraphics[scale = 0.27] {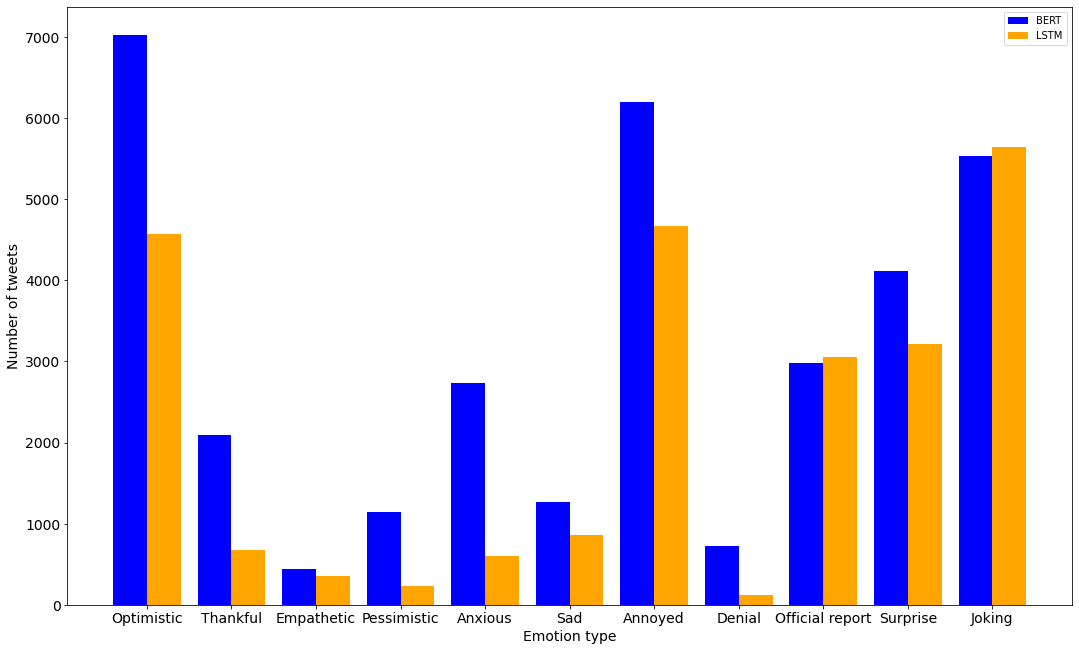}
 }  
 
 \subfigure[Delhi]{
   \includegraphics[scale = 0.27] {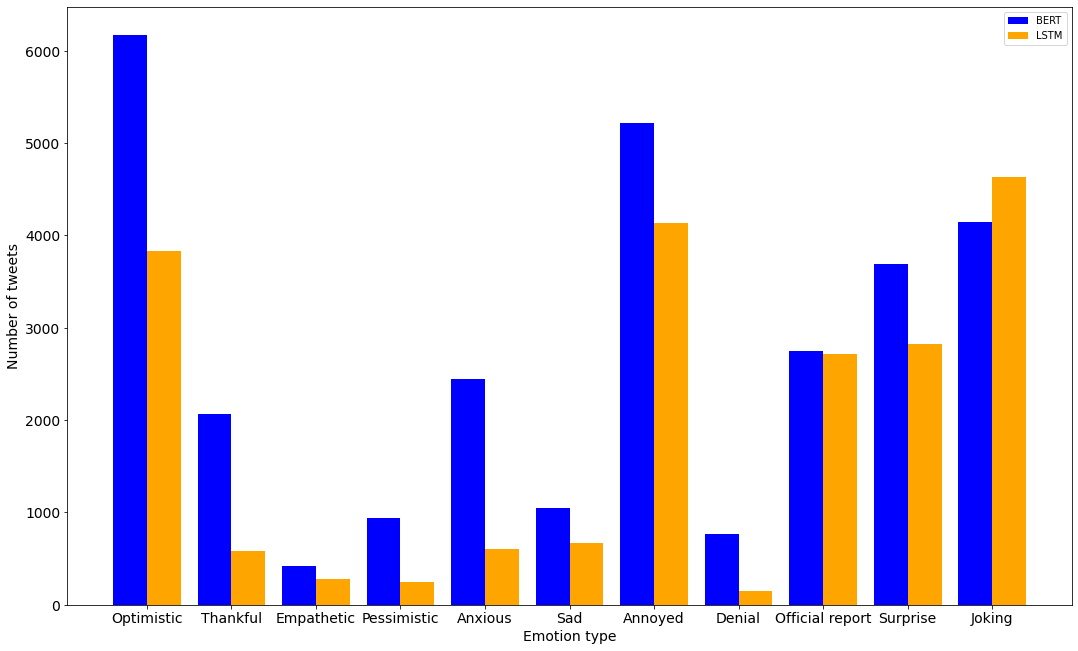}
 }  
\caption{Distribution of sentiments predicted for the respective datasets by the \textcolor{black}{LSTM and BERT models.}}

\label{fig:senti_distri}
\end{figure}

In this section, we present results of the predictions of the COVID-19 tweets in India, Maharashtra, and Delhi treated as separate datasets. We note that in our Senwave dataset  analysed in previous section has been used for training. The label ``official report", which is not a sentiment was included in the Senwave dataset, and hence we   use it  for prediction. Figure \ref{fig:senti_distri} presents distribution of sentiments predicted  by the  LSTM and BERT models for the respective datasets for the entire time-span of interest (March to September 2020). In the case of India (Panel a), we notice that the ``optimistic", ``annoyed" and ``joking" sentiments are the most expressed which also follows in the case of Maharashtra and Delhi (Panel b and c). We notice that the BERT model seems to capture more sentiments expressed when compared to the LSTM model, particularly  the ``optimistic", ``anxious",  and ``annoyed" sentiments.   We find that negative sentiments such as ``pessimistic", ``anxious" and ``sad" have been least expressed. We find ``optimistic" sentiment as the  most prominent sentiment which is followed by ``annoyed" and ``joking". It is simpler to label "optimistic" and "thankful" as a positive sentiment, but it becomes increasing difficult when it comes to the sentiments ``joking" and ``surprise", when context information is not available. Hence, it is insightful to review the sentiments such as ``joking" and ``surprise" in relation to other sentiments expressed.   We find a similar trend in the case of Delhi and Maharashtra (Panel b and c) which are subsets of data from India (Panel a), the only major difference is the number of tweets which is significantly lower due to their respective  population.  Since the  BERT model provided the best results for the training data, we provide the results  by the BERT model henceforward.

\begin{figure}[htbp!]
\includegraphics[width=10cm]{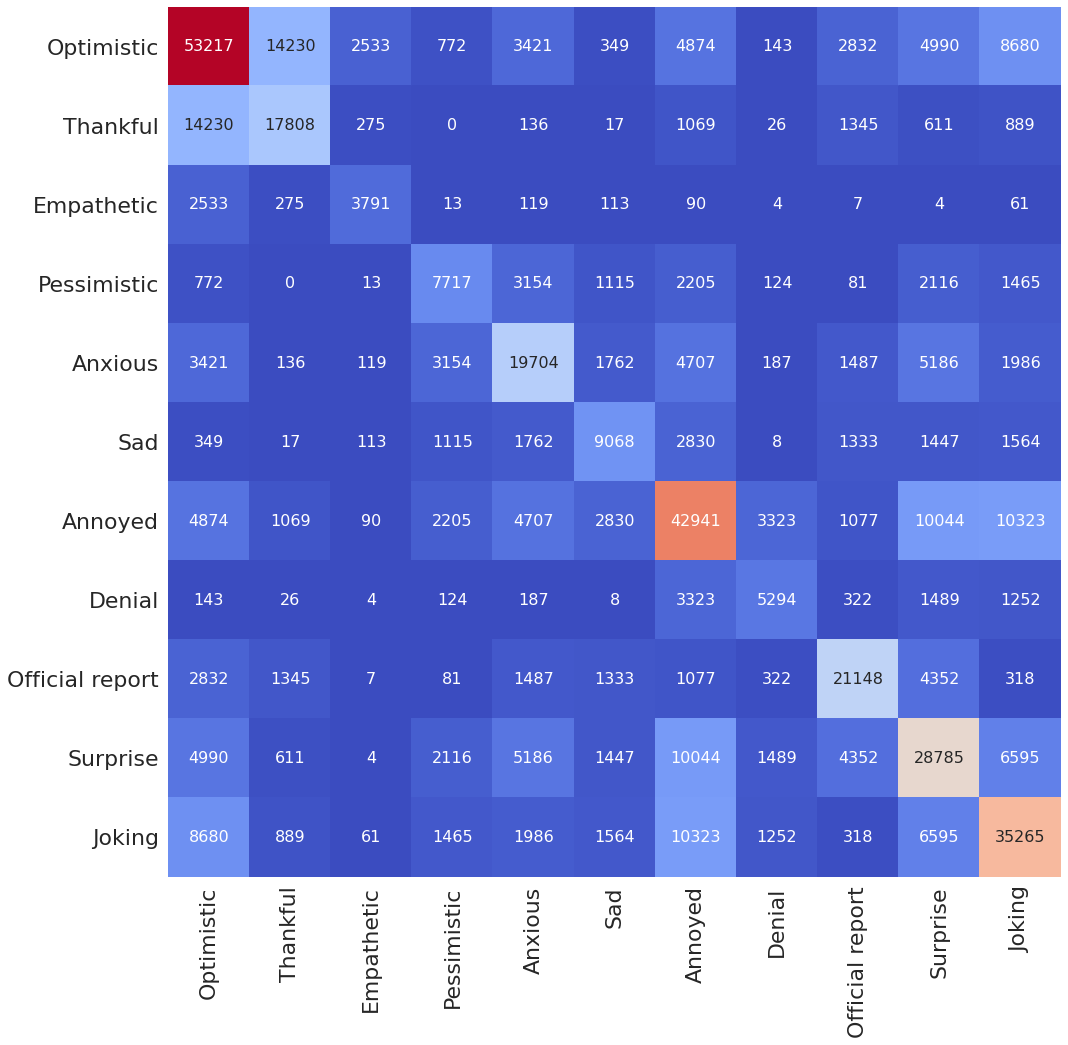}
\caption{Heatmap  showing number of occurrence  of a given sentiment in relation to the rest of the sentiments for the India dataset   \textcolor{black}{using the BERT model.}}
\label{fig:heatoccur}
\end{figure}

\begin{figure}[htbp!]
\includegraphics[width=10cm]{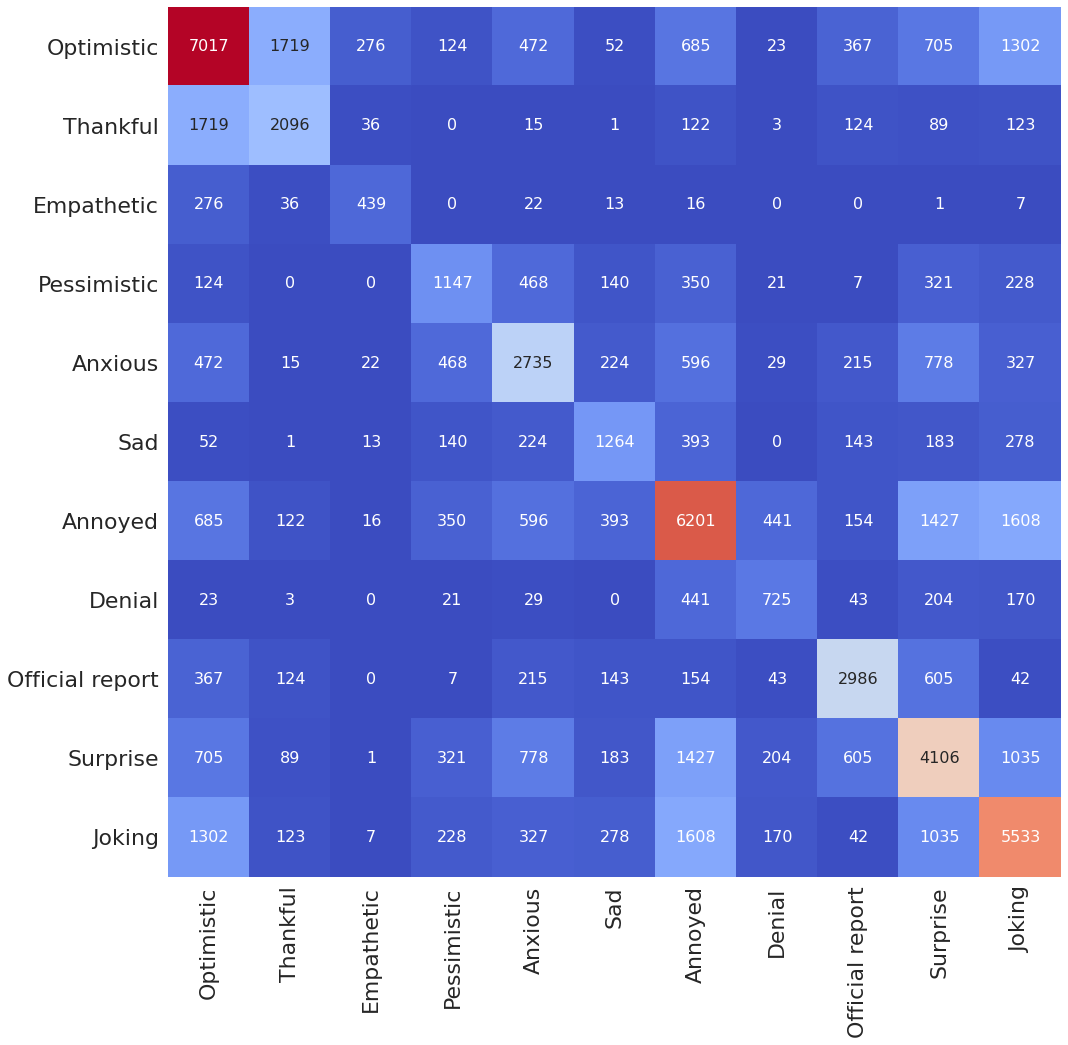}
\caption{Heatmap  showing number of occurrence  of a given sentiment in relation to the rest of the sentiments for the Maharashtra dataset  \textcolor{black}{using the BERT model.}}
\label{fig:heatoccur_maharastra}
\end{figure}

\begin{figure}[htbp!]
\includegraphics[width=10cm]{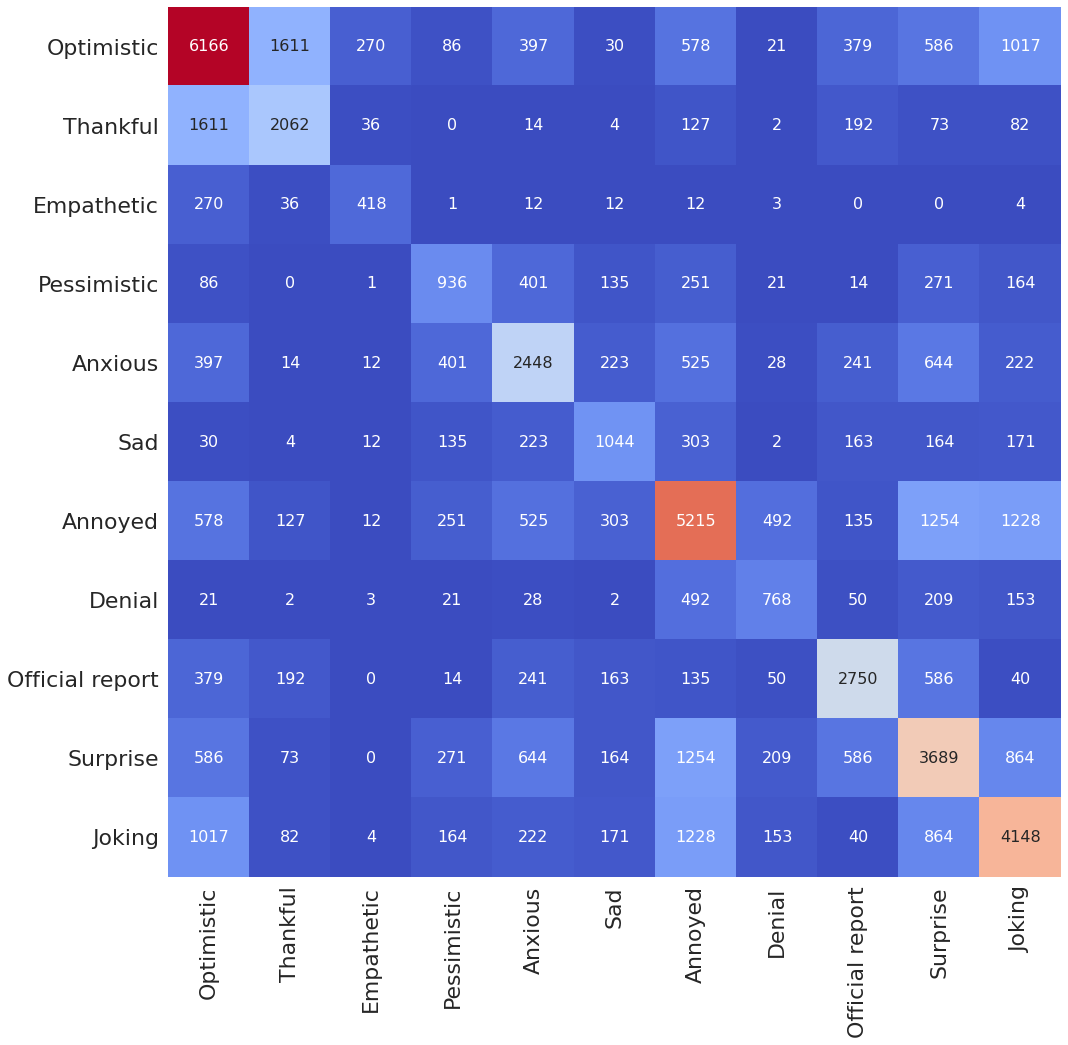}
\caption{Heatmap  showing number of occurrence  of a given sentiment in relation to the rest of the sentiments for the Delhi dataset \textcolor{black}{using the BERT model.}}
\label{fig:heatoccur_delhi}
\end{figure}

Next, we review the sentiments predicted  using a heatmap to examine number of occurrence  of a given sentiment in relation to the rest of the sentiments (Figures \ref{fig:heatoccur}, 7 and 8). These heatmaps essentially indicate how two sentiments have been expressed at once and provides more insights regarding the context of the negative and positive sentiments given in Figure \ref{fig:senti_distri}. Note that in  Figure \ref{fig:senti_distri}, we found that the third most prominent sentiment has been ``joking"; however, we were not sure if it is positive or negative sentiment. As shown in Figure \ref{fig:heatoccur}, \textcolor{black}{we find that most tweets that are associated with ``joking" are either ``optimistic" (8680) or ``annoyed" (10323), and some are also  ``thankful" (889). A much lower portion (below 500) are either ``empathetic" or ``official report" while ``joking". Next, we review the sentiment ``optimistic" and find that majority are either ``thankful" (14230) or ``joking" (8680). The  ``optimistic" tweets  with negative sentiments are relatively minor, such as ``pessimistic" (772), ``denial" (143), and ``sad" (349). Furthermore, a significant portion are  also ``empathetic" (2533).} It is not common for one to make a statement that is optimistic and pessimistic at the same time; hence, this could be a   wrong prediction by the model. However, it seems the model is making the right prediction when we look at the heatmap for the hand-labelled training datasets (Figure \ref{fig:heatoccur_senwave}), where such combinations of sentiments  have been labelled by experts.  We show examples of tweets of this case and compare  with those that are optimistic and thankful in Table \ref{tab:optipessimis}.

  Figure \ref{fig:heatoccur_maharastra} and Figure \ref{fig:heatoccur_delhi} show the number of occurrence  of a given sentiment in relation to the rest of the sentiments for the case of Maharashtra and Delhi, which follows a similar pattern when compared to case of India (Figure  \ref{fig:heatoccur})). We note that the Senwave dataset which shows tweets worldwide (Figure \ref{fig:heatoccur_senwave}) follow a similar pattern than the case in Indian datasets when it comes to sentiments such as ``joking" and being ``optimistic" or ``joking" and ``annoyed". Senwave dataset also features cases of being optimistic and pessimistic at the same time (226 cases). This could be due to sentiments expressed in two opposing sentences in a tweet.   We can infer that since such patterns are part of the training data, it cannot be an error in predictions when looking at the Indian datasets.

    \begin{table*}[htbp!]
        \small
       \centering
        \begin{tabular}{l l l} 
        \hline
        Month & Tweet &  Sentiment combination\\
        \hline
         \hline
        May & ``don’t care china only care indian covid19 news" & ``optimistic - pessimistic"  \\
        \hline
        August & ``dear uddhav ji other than covid problem, & ``optimistic - pessimistic" \\
        
         &   what ever wrong is happening is not good   &   \\
             &     for your government" &   \\
        \hline
        September & ``sir, plz look in have benefits to sr citizen." & ``optimistic - pessimistic" \\
        \hline
        April & ``thank you switzerland smiling face with smiling   & ``optimistic - thankful" \\
            &   eyes especially  zermatt for showing   " &   \\
                &      solidarity for india flag" &   \\
        \hline
        May & ``big thanks to the cfpc....thoroughly enjoyed it!"  & "optimistic - thankful" \\
        \hline
        June & ``doctors, activists urge pm to promote plant-based diet & ``optimistic - thankful"  \\
        
          &   |   india news - times of india" &  \\
        \hline
         \hline
        
        \end{tabular}
        \caption{Selected example   that show cases of tweets that are  ``optimistic" and also ``pessimistic", along with cases that are ``optimistic" and also ``thankful".}
        \label{tab:optipessimis}
    \end{table*}

\begin{figure}[htbp!]
\centering
\subfigure[Senwave hand-labelled sentiments (Worldwide)]{
\includegraphics[scale =0.25]{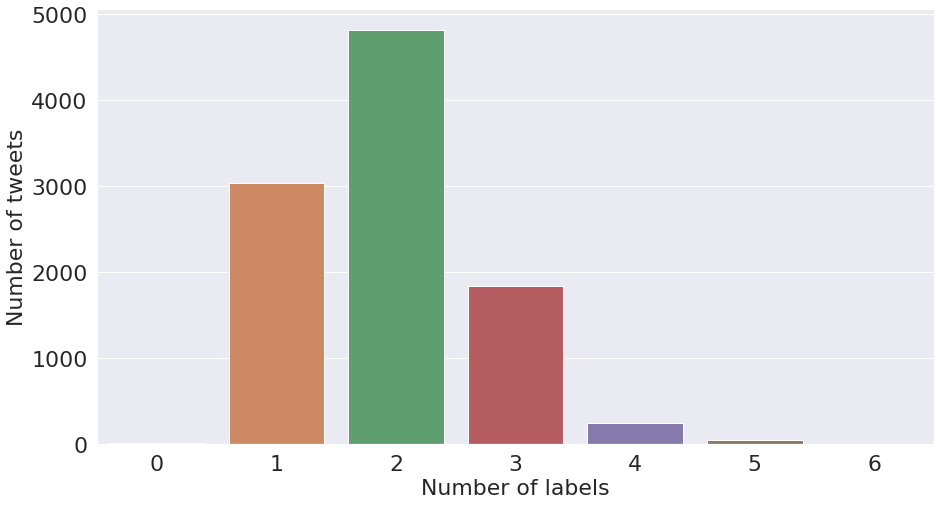}
 }
 \subfigure[Predicted (India) \textcolor{black}{using BERT}]{
 \includegraphics[scale =0.25]{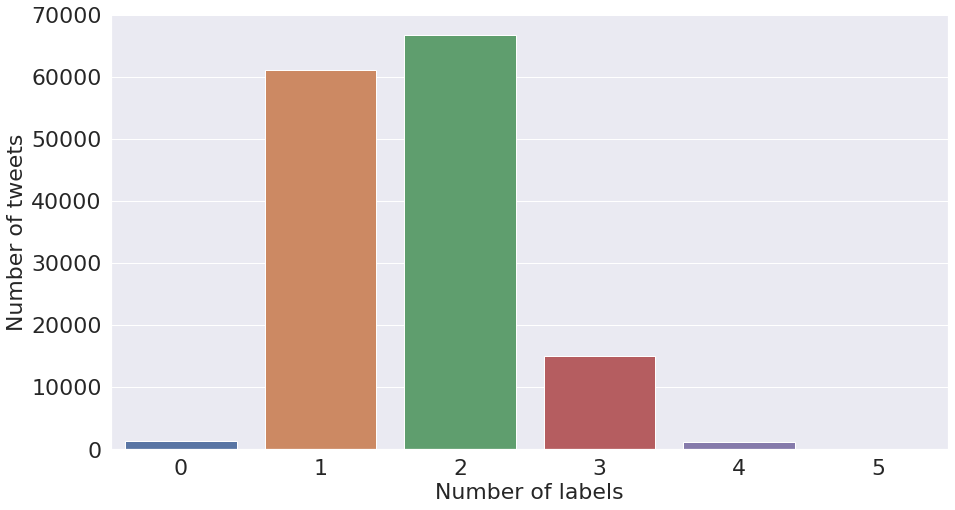}
 }  
 \subfigure[Predicted (India) \textcolor{black}{using LSTM}]{
 \includegraphics[scale =0.25]{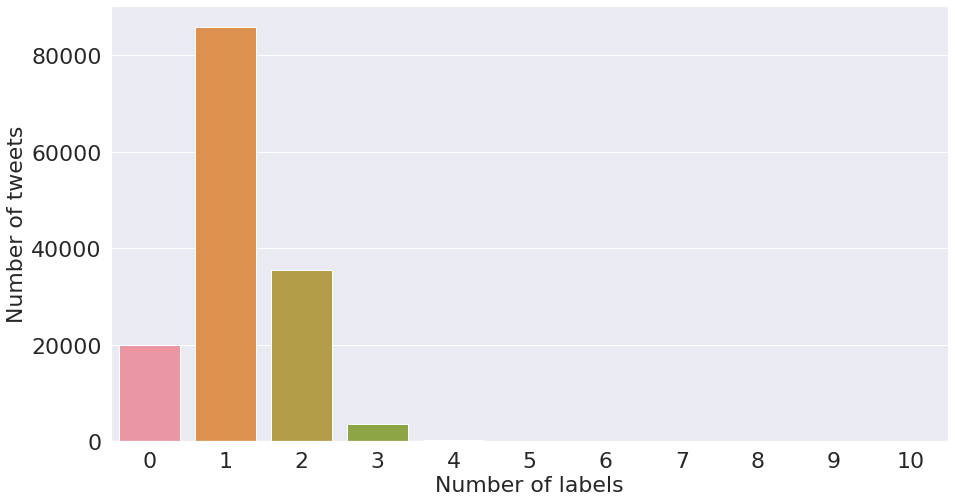}
 }
   
\caption{Distribution of tweets with number of  combination of sentiment types (labels) \textcolor{black}{for Senwave training dataset, and predictions by BERT and LSTM models.}}

\label{fig:numbsenti}
\end{figure}

Figure \ref{fig:numbsenti} provides a visualisation of the distribution of tweets with  number of combination  sentiments. We find that  around 60\% of the tweets have a singular sentiment. Moreover, about  25\% of the tweets have two  sentiment attached them and 14\% have no  sentiment  attached to them. Furthermore, a small number of tweets have 3 or more emotions attached to them which indicates that  much often, people do not show multiple emotions at the same time.

\begin{figure*}[htbp!]
\centering
\includegraphics[width=14cm]{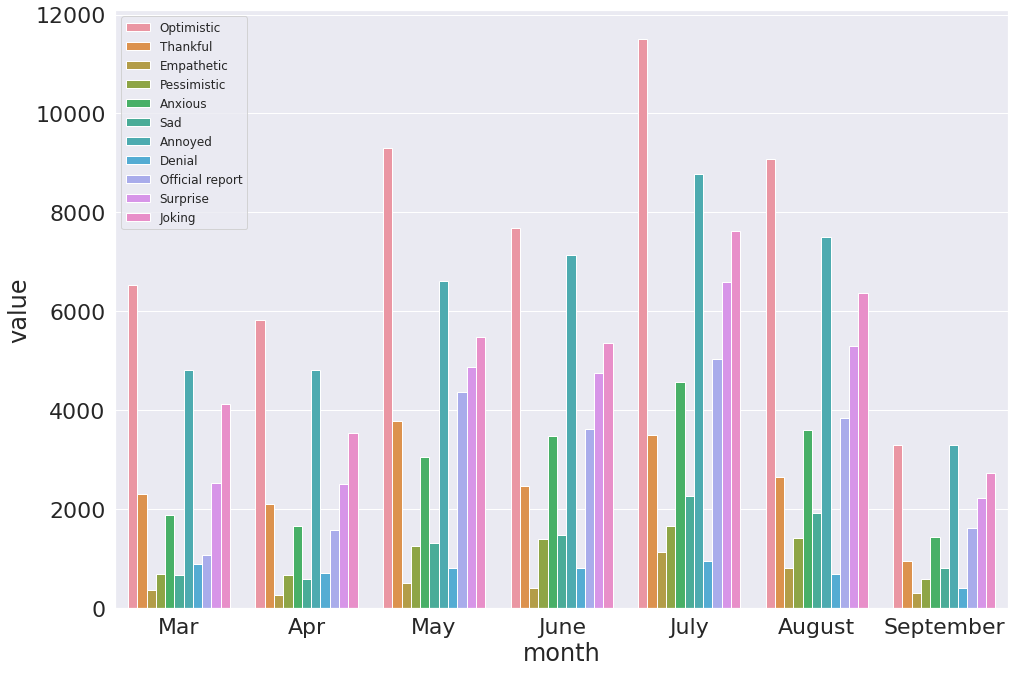}
\caption{Monthly COVID-19  sentiments in India \textcolor{black}{using the BERT model.}}
\label{fig:monindia}
\end{figure*}

\begin{figure*}[htbp!]
\centering
\includegraphics[width=14cm]{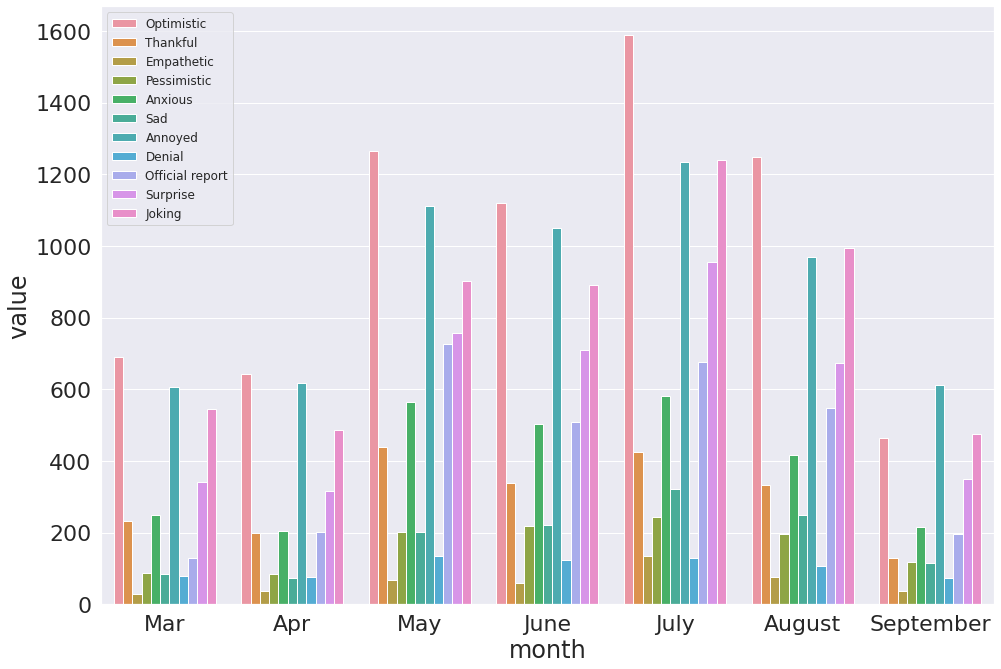}
\caption{Monthly COVID-19  sentiments in Maharashtra \textcolor{black}{using the BERT model.}}
\label{fig:monmaha}
\end{figure*}

\begin{figure*}[htbp!]
\centering
\includegraphics[width=14cm]{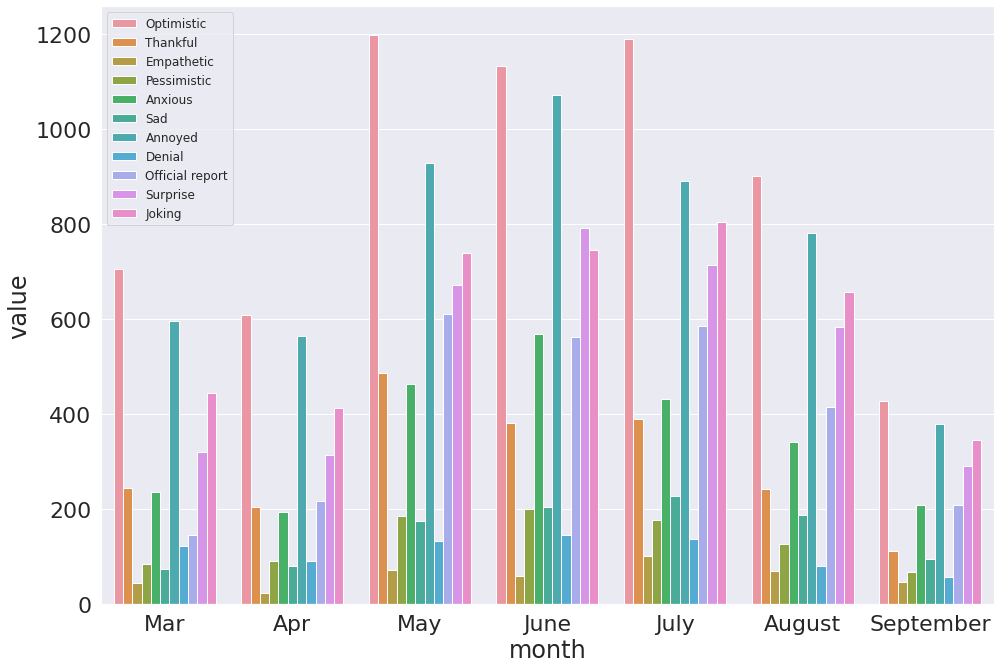}
\caption{Monthly COVID-19  sentiments in Delhi \textcolor{black}{using the BERT model.}}
\label{fig:mondelhi}
\end{figure*}

Finally, we present results that visualises the trend of sentiments expressed over time in the three datasets. This is one of the key findings which can enable a understanding of the reaction in terms of emotions expressed by the population given rise and fall of COVID-19 novel cases as shown in Figure \ref{fig:cases}. 
Figures \ref{fig:monindia}, \ref{fig:monmaha} and \ref{fig:mondelhi} presents visualisation  of the monthly sentiments of India, Maharashtra and Delhi, respectively.

 \section{Discussion}

Humans can express more than one sentiment at a time; however,  there are variations in the number of sentiments that can be expressed by facial expressions \cite{jack2014dynamic}, when compared to spoken  or written emotions  such as  tweets. As shown in Figure 
\ref{fig:numbsenti}, majority of the tweets have one sentiment, both in hand-labelled (Panel a) and predicted datasets (Panel b). This is followed by two sentiments, while a minority have 3 sentiments. We note that there is a significant portion of tweets  with no sentiments and there are no tweets of more than three sentiments at a time. The study of emotions with technology has gained much interested in last decade which gave much progress in understanding sentiments.  Trampe et al. \cite{trampe2015emotions} presented a study of everyday emotional experiences through an experience sampling smartphone application  that monitored real-time emotions of more than  11,000  participants and found the group experienced at least one emotion 90\% of the time with joy as the  the most frequent emotion, followed by love and anxiety. The type of emotion would be highly dependent on the study region which featured Europeans with majority  French participants. Cowen and Keltner reported \cite{cowen2017self}    twenty seven distinct categories of emotions bridged by continuous gradients by  data from  emotionally evocative short videos with   varying situational content. These studies contributed to better understanding of emotions given historical viewpoints about context and definitions of emotions and associations between them \cite{wintre1994developmental,kemper1987many,ekman1992argument,lucas2003measuring}. However, we did not find any study that reviews the number of emotions that can be expressed at a time in relation to catastrophic events that keep changing with time, such as the rise of COVID-19 cases. 

We revisit the case of Indian dataset (Figure \ref{fig:monindia}), where the monthly tweets did not sharply increase with number of novel cases (Figure 2) with a nationwide peak of novel cases (Figure 2, Panel a). The number of tweets gradually increased with peak of tweets in July (Figure 2, Panel a). When India had a peak of novel cases, we found that the number of tweets significantly lowered. Hence, people have been alarmed by rising cases, but a month before the peak of novel cases was seen, the  tweets were reduced. Moreover,  we find that the ``optimistic", ``annoyed" and ``joking" tweets are mostly dominating the monthly tweets for India (Figure 10) and Maharashtra (Figure 11), with a mix of annoyed sentiments in case of Delhi (Figure 12). There is significantly lower number of negative sentiments for the respective datasets (Figure 10-12).  

We note that a limitation of the framework is due to the Senwave training data which considered tweets worldwide during COVID-19 by a group of experts; however, there can be limitations on how experts  perceive  humour in different regions.  Humour is expressed differently in different cultural groups, i.e. a tweet that may be humorous  in USA may not be taken as humorous in  India due to cultural and language dimensions.  There are several studies about the effect of humour in changing cultural or regional settings \cite{friedman2014comedy,rossato2010audiences,kuipers2011politics}. A good example is that in orthodox Chinese literature, humour was not expressed due to religious taboo in ancient  Buddhism which later eased with Zen Buddhism \cite{chey2011youmo,clasquin2001real}; however, the Hindu literature had a different or eased attitude towards humour as expressed in ancient Hindu texts \cite{elst2011humour}. Although historic textual composition cannot be related to how tweets are expressed in India, it is important to note the cultural and language differences in how humour has been expressed. 

 Another  limitation of the study is regarding the uncertainty of the data. Our results are based on the tweets expressed; however, we note that  only a small fraction of the population generally express their views on Twitter. Moreover, we need to be aware that due to restrictions on freedom of speech and political biases, not everyone can express themselves freely. Social media networks have been active in monitoring how people express themselves in order to limit the rise of anti-vaccine sentiments. Furthermore, our framework’s training data is based on  tweets expressed worldwide, whereas the test data is based on tweets from India during the pandemic. There is a large language diversity in India and at times, people express themselves with a combination of languages with  geographic and culture specific jargon. Although, we convert them to English, certain limitations in capturing the context will exist.

We note that there has not been much work in uncertainty quantification for the predictions and there are different level of uncertainties, particularly in model parameters and data. We note that the training data is hand-labelled, and at times two or three sentiments have been expressed at once. This could be something open to interpretation by experts as it is hard to formally detect more than one sentiment from a tweet of only thirty words. Hence, the expert labelled training dataset adds to uncertainty in model predictions.  In future work, Bayesian deep learning can provide  a way to address uncertainty in model predictions \cite{chandra2019langevin,
chandra2021revisiting,chandra2021bayesian}.

\section{Conclusions}

We presented a study with novel deep learning models for sentimental analysis  during the rise of  COVID-19 infections. We selected tweets from India as our case study and reviewed tweets from specific regions that included Maharastra and Delhi. We took advantage of COVID-19 dataset of   10,000 hand-labelled  tweets   for training the respective deep learning models. Our investigation revealed that   majority of the tweets have been  ``optimistic", ``annoyed"  and ``joking"  that expresses optimism, fear and uncertainty  during the rise of the COVID-19 cases in India.    The number of tweets significantly lowered towards the peak of new cases. Furthermore,  the optimistic, annoyed and joking tweets   mostly dominated the monthly tweets with much lower number of negative sentiments expressed. We found that most tweets that have been associated with ``joking" were either ``optimistic"  or ``annoyed”, and minority of them were  also ``thankful".  In terms of the ``annoyed” sentiments in tweets, mostly were either ``surprised" or ``joking". 
These predictions generally  indicate that although the majority have been optimistic, a significant  group of population has been annoyed towards the way the pandemic was handled by the authorities. The major contribution of the paper is the framework which provides sentiment analysis in a population given the rise of the COVID-19 cases. The framework   can be used by officials for   better COVID-19 management through  policies and projects, such as   support  for depression and mental health issues.

Future work can use the framework for different regions, countries, ethnic and social groups to understand their behaviour given   multiple peaks of novel cases. The framework can be extended to understand reactions towards vaccinations with the rise of anti-vaccine sentiments   given fear, insecurity and unpredictability of COVID-19.  Finally, the framework can incorporate topic modelling with sentiment analysis which can provide more details   for emerging topics during the rise of COVID-19 cases in relation to  various government protocols such as  lock-downs and vaccination plans.

\section*{Data and Code}

We provide Python-based open source code and data for further research
(\url{https://github.com/sydney-machine-learning/COVID19_sentinentanalysis}).


\begin{thebibliography}{10}

\bibitem{golbeck2011predicting}
Golbeck J, Robles C, Edmondson M, Turner K.
\newblock Predicting personality from twitter.
\newblock In: 2011 IEEE third international conference on privacy, security,
  risk and trust and 2011 IEEE third international conference on social
  computing. IEEE; 2011. p. 149--156.

\bibitem{quercia2011our}
Quercia D, Kosinski M, Stillwell D, Crowcroft J.
\newblock Our twitter profiles, our selves: Predicting personality with
  twitter.
\newblock In: 2011 IEEE third international conference on privacy, security,
  risk and trust and 2011 IEEE third international conference on social
  computing. IEEE; 2011. p. 180--185.

\bibitem{bittermann2021mining}
Bittermann A, Batzdorfer V, M{\"u}ller SM, Steinmetz H.
\newblock Mining Twitter to detect hotspots in psychology.
\newblock Zeitschrift f{\"u}r Psychologie. 2021;.

\bibitem{lin2015building}
Lin J.
\newblock On building better mousetraps and understanding the human condition:
  Reflections on big data in the social sciences.
\newblock The ANNALS of the American Academy of Political and Social Science.
  2015;659(1):33--47.

\bibitem{coppersmith2014quantifying}
Coppersmith G, Dredze M, Harman C.
\newblock Quantifying mental health signals in Twitter.
\newblock In: Proceedings of the workshop on computational linguistics and
  clinical psychology: From linguistic signal to clinical reality; 2014. p.
  51--60.

\bibitem{murphy2017hands}
Murphy SC.
\newblock A hands-on guide to conducting psychological research on Twitter.
\newblock Social Psychological and Personality Science. 2017;8(4):396--412.

\bibitem{zhou2019comparative}
Zhou Y, Na JC.
\newblock A comparative analysis of Twitter users who Tweeted on psychology and
  political science journal articles.
\newblock Online Information Review. 2019;43(7):1188-1208.

\bibitem{wang2016twitter}
Wang W, Hernandez I, Newman DA, He J, Bian J.
\newblock Twitter analysis: Studying US weekly trends in work stress and
  emotion.
\newblock Applied Psychology. 2016;65(2):355--378.

\bibitem{alizadeh2019psychology}
Alizadeh M, Weber I, Cioffi-Revilla C, Fortunato S, Macy M.
\newblock Psychology and morality of political extremists: evidence from
  Twitter language analysis of alt-right and Antifa.
\newblock EPJ Data Science. 2019;8(17): \newblock https://doi.org/10.1140/epjds/s13688-019-0193-9

\bibitem{garg2017sentiment}
Garg P, Garg H, Ranga V.
\newblock Sentiment analysis of the Uri terror attack using Twitter.
\newblock In: 2017 International conference on computing, communication and
  automation (ICCCA). IEEE; 2017. p. 17--20.

\bibitem{manning1999foundations}
Manning C, Schutze H.
\newblock Foundations of statistical natural language processing.
\newblock MIT press; 1999.

\bibitem{liu2012survey}
Liu B, Zhang L.
\newblock A survey of opinion mining and sentiment analysis.
\newblock In: Mining text data. Springer; 2012. p. 415--463.

\bibitem{medhat2014sentiment}
Medhat W, Hassan A, Korashy H.
\newblock Sentiment analysis algorithms and applications: A survey.
\newblock Ain Shams engineering journal. 2014;5(4):1093--1113.

\bibitem{hussein2018survey}
Hussein DMEDM.
\newblock A survey on sentiment analysis challenges.
\newblock Journal of King Saud University-Engineering Sciences.
  2018;30(4):330--338.

\bibitem{ordenes2014analyzing}
Ordenes FV, Theodoulidis B, Burton J, Gruber T, Zaki M.
\newblock Analyzing customer experience feedback using text mining: A
  linguistics-based approach.
\newblock Journal of Service Research. 2014;17(3):278--295.

\bibitem{greaves2013use}
Greaves F, Ramirez-Cano D, Millett C, Darzi A, Donaldson L.
\newblock Use of sentiment analysis for capturing patient experience from
  free-text comments posted online.
\newblock Journal of medical Internet research. 2013;15(11):e239.

\bibitem{mittal2012stock}
Mittal A, Goel A.
\newblock Stock prediction using Twitter sentiment analysis.
\newblock Standford University, CS229 (2011 http://cs229 stanford
  edu/proj2011/GoelMittal-StockMarketPredictionUsingTwitterSentimentAnalysis
  pdf). 2012;15.

\bibitem{wang2012system}
Wang H, Can D, Kazemzadeh A, Bar F, Narayanan S.
\newblock A system for real-time Twitter sentiment analysis of 2012 us
  presidential election cycle.
\newblock In: Proceedings of the ACL 2012 system demonstrations; 2012. p.
  115--120.

\bibitem{zhang2018deep}
Zhang L, Wang S, Liu B.
\newblock Deep learning for sentiment analysis: A survey.
\newblock Wiley Interdisciplinary Reviews: Data Mining and Knowledge Discovery.
  2018;8(4):e1253.

\bibitem{kouloumpis2011twitter}
Kouloumpis E, Wilson T, Moore J.
\newblock Twitter sentiment analysis: The good the bad and the OMG!
\newblock In: Proceedings of the Fifth International Conference on Weblogs and Social Media;  2011. p. 538-541

\bibitem{giachanou2016like}
Giachanou A, Crestani F.
\newblock Like it or not: A survey of Twitter sentiment analysis methods.
\newblock ACM Computing Surveys. 2016;49(2):1--41.

\bibitem{severyn2015twitter}
Severyn A, Moschitti A.
\newblock Twitter sentiment analysis with deep convolutional neural networks.
\newblock In: Proceedings of the 38th international ACM SIGIR conference on
  research and development in information retrieval; 2015. p. 959--962.

\bibitem{Gorbalenya2020species}
Gorbalenya AE, Baker SC, Baric RS, de~Groot RJ, Drosten C, Gulyaeva AA, et~al.
\newblock The species Severe acute respiratory syndrome-related coronavirus:
  classifying 2019-nCoV and naming it SARS-CoV-2.
\newblock Nature Microbiology. 2020;5:536–544.

\bibitem{monteil2020inhibition}
Monteil V, Kwon H, Prado P, Hagelkr{\"u}ys A, Wimmer RA, Stahl M, et~al536–544
\newblock Inhibition of SARS-CoV-2 infections in engineered human tissues using
  clinical-grade soluble human ACE2.
\newblock Cell.  2020;181: 905--913

\bibitem{world2020coronavirus}
Organization WH, et~al.. Coronavirus disease 2019 ({COVID-19}): situation
  report, 72; 2020.
\newblock Available from: \url{https://apps.who.int/iris/handle/10665/331685}.

\bibitem{cucinotta2020declares}
Cucinotta D, Vanelli M.
\newblock {WHO} declares COVID-19 a pandemic.
\newblock Acta bio-medica: Atenei Parmensis. 2020;91(1):157--160.

\bibitem{siche2020impact}
Siche R.
\newblock What is the impact of {COVID-19} disease on agriculture?
\newblock Scientia Agropecuaria. 2020;11(1):3--6.

\bibitem{richards2020impact}
Richards M, Anderson M, Carter P, Ebert BL, Mossialos E.
\newblock The impact of the COVID-19 pandemic on cancer care.
\newblock Nature Cancer. 2020;1(6):565--567.

\bibitem{tiwari2021delhi}
Tiwari A, Gupta R, Chandra R.
\newblock Delhi air quality prediction using LSTM deep learning models with a
  focus on {COVID-19} lockdown.
\newblock arXiv preprint arXiv:210210551. 2021.

\bibitem{shinde2020forecasting}
Shinde GR, Kalamkar AB, Mahalle PN, Dey N, Chaki J, Hassanien AE.
\newblock Forecasting models for coronavirus disease (COVID-19): a survey of
  the state-of-the-art.
\newblock SN Computer Science. 2020;1(4):1--15.

\bibitem{rahimi2021review}
Rahimi I, Chen F, Gandomi AH.
\newblock A review on COVID-19 forecasting models.
\newblock Neural Computing and Applications. 2021; p. 1--11.
\newblock https://doi.org/10.1007/s00521-020-05626-8

\bibitem{zeroual2020deep}
Zeroual A, Harrou F, Dairi A, Sun Y.
\newblock Deep learning methods for forecasting COVID-19 time-Series data: A
  Comparative study.
\newblock Chaos, Solitons \& Fractals. 2020;140:110121.

\bibitem{ChandraCOVID2021}
Chandra R, Jain A, Chauhan DS.
\newblock Deep learning via LSTM models for {COVID-19} infection forecasting
  in India.
\newblock CoRR. 2021;abs/2101.11881.

\bibitem{TiwariAir2021}
Tiwari A, Gupta R, Chandra R.
\newblock Delhi air quality prediction using {LSTM} deep learning models with a
  focus on COVID-19 lockdown.
\newblock CoRR. 2021;abs/2102.10551.

\bibitem{chakraborty2020sentiment}
Chakraborty K, Bhatia S, Bhattacharyya S, Platos J, Bag R, Hassanien AE.
\newblock Sentiment Analysis of COVID-19 tweets by Deep Learning
  Classifiers—A study to show how popularity is affecting accuracy in social
  media.
\newblock Applied Soft Computing. 2020;97:106754.

\bibitem{barkur2020sentiment}
Barkur G, Vibha GBK.
\newblock Sentiment analysis of nationwide lockdown due to COVID 19 outbreak:
  Evidence from India.
\newblock Asian journal of psychiatry. 2020;51:102089.

\bibitem{abd2020top}
Abd-Alrazaq A, Alhuwail D, Househ M, Hamdi M, Shah Z.
\newblock Top concerns of tweeters during the COVID-19 pandemic: infoveillance
  study.
\newblock Journal of medical Internet research. 2020;22(4):e19016.

\bibitem{xue2020public}
Xue J, Chen J, Chen C, Zheng C, Li S, Zhu T.
\newblock Public discourse and sentiment during the COVID 19 pandemic: Using
  Latent Dirichlet Allocation for topic modeling on Twitter.
\newblock PloS one. 2020;15(9):e0239441.

\bibitem{hung2020social}
Hung M, Lauren E, Hon ES, Birmingham WC, Xu J, Su S, et~al.
\newblock Social network analysis of COVID-19 Sentiments: application of
  artificial intelligence.
\newblock Journal of medical Internet research. 2020;22(8):e22590.

\bibitem{wang2020covid}
Wang T, Lu K, Chow KP, Zhu Q.
\newblock COVID-19 Sensing: Negative sentiment analysis on social media in
  China via Bert Model.
\newblock Ieee Access. 2020;8:138162--138169.

\bibitem{zhou2020examination}
Zhou J, Yang S, Xiao C, Chen F.
\newblock Examination of community sentiment dynamics due to COVID-19 pandemic:
  A case study from Australia.
\newblock arXiv preprint arXiv:200612185. 2020;.


 

\bibitem{pokharel2020twitter}
Pokharel BP.
\newblock Twitter sentiment analysis during COVID-19 outbreak in Nepal.
\newblock Available at SSRN 3624719. 2020.

\bibitem{de2020sentiment}
de~Las Heras-Pedrosa C, S{\'a}nchez-N{\'u}{\~n}ez P, Pel{\'a}ez JI.
\newblock Sentiment analysis and emotion understanding during the COVID-19
  pandemic in Spain and its impact on digital ecosystems.
\newblock International Journal of Environmental Research and Public Health.
  2020;17(15):5542;
  \newblock https://doi.org/10.3390/ijerph17155542

\bibitem{kruspe2020cross}
Kruspe A, H{\"a}berle M, Kuhn I, Zhu XX.
\newblock Cross-language sentiment analysis of European Twitter messages
  during the COVID-19 pandemic.
\newblock arXiv preprint arXiv:200812172. 2020.

\bibitem{yang2020senwave}
Yang Q, Alamro H, Albaradei S, Salhi A, Lv X, Ma C, et~al.
\newblock SenWave: Monitoring the Global Sentiments under the COVID-19
  Pandemic.
\newblock arXiv preprint arXiv:200610842. 2020.

\bibitem{li2018word}
Li Y, Yang T.
\newblock Word embedding for understanding natural language: a survey.
\newblock In: Guide to big data applications. Springer; 2018. p. 83--104.

\bibitem{kutuzov2018diachronic}
Kutuzov A, {\O}vrelid L, Szymanski T, Velldal E.
\newblock Diachronic word embeddings and semantic shifts: a survey.
\newblock arXiv preprint arXiv:180603537. 2018;.

\bibitem{ruder2019survey}
Ruder S, Vuli{\'c} I, S{\o}gaard A.
\newblock A survey of cross-lingual word embedding models.
\newblock Journal of Artificial Intelligence Research. 2019;65:569--631.

\bibitem{zhang2010understanding}
Zhang Y, Jin R, Zhou ZH.
\newblock Understanding bag-of-words model: a statistical framework.
\newblock International Journal of Machine Learning and Cybernetics.
  2010;1(1-4):43--52.

\bibitem{ramos2003using}
Ramos J, et~al.
\newblock Using TF-IFD to determine word relevance in document queries.
\newblock In: Proceedings of the first instructional conference on machine
  learning. vol. 242.; 2003. p. 29--48.

\bibitem{goodman2001bit}
Goodman JT.
\newblock A bit of progress in language modeling.
\newblock Computer Speech \& Language. 2001;15(4):403--434.

\bibitem{guthrie2006closer}
Guthrie D, Allison B, Liu W, Guthrie L, Wilks Y.
\newblock A closer look at skip-gram modelling.
\newblock In: LREC. vol.~6.; 2006. p. 1222--1225.

\bibitem{mikolov2013distributed}
Mikolov T, Sutskever I, Chen K, Corrado G, Dean J.
\newblock Distributed representations of words and phrases and their
  compositionality.
\newblock arXiv preprint arXiv:13104546. 2013.

\bibitem{pennington2014glove}
Pennington J, Socher R, Manning CD.
\newblock GloVe: Global Vectors for Word Representation.
\newblock In: Empirical Methods in Natural Language Processing (EMNLP); 2014.
  p. 1532--1543. 

\bibitem{zhao2018learning}
Zhao J, Zhou Y, Li Z, Wang W, Chang KW.
\newblock Learning gender-neutral word embeddings.
\newblock arXiv preprint arXiv:180901496. 2018.

\bibitem{ghannay2016evaluation}
Ghannay S, Esteve Y, Camelin N, Del{\'e}glise P.
\newblock Evaluation of acoustic word embeddings.
\newblock In: Proceedings of the 1st Workshop on Evaluating Vector-Space
  Representations for NLP; 2016. p. 62--66.

\bibitem{schwenk2007continuous}
Schwenk H.
\newblock Continuous space language models.
\newblock Computer Speech \& Language. 2007;21(3):492--518.

\bibitem{wang2018comparison}
Wang Y, Liu S, Afzal N, Rastegar-Mojarad M, Wang L, Shen F, et~al.
\newblock A comparison of word embeddings for the biomedical natural language
  processing.
\newblock Journal of biomedical informatics. 2018;87:12--20.

\bibitem{elman_Zipser1988}
Elman JL, Zipser D.
\newblock Learning the hidden structure of speech.
\newblock The Journal of the Acoustical Society of America.
  1988;83(4):1615--1626. 

\bibitem{Elman_1990}
Elman JL.
\newblock Finding structure in time.
\newblock Cognitive Science. 1990;14:179--211.

\bibitem{Omlin_etal1996}
Omlin CW, Giles CL.
\newblock Constructing deterministic finite-state automata in recurrent neural
  networks.
\newblock J ACM. 1996;43(6):937--972. 

\bibitem{Omlin_Giles1992}
Omlin CW, Giles CL.
\newblock Training second-order recurrent neural networks using hints.
\newblock In: Proceedings of the Ninth International Conference on Machine
  Learning. Morgan Kaufmann; 1992. p. 363--368.

\bibitem{Chandra_Omlin2007}
Chandra R, Omlin CW.
\newblock The Comparison and Combination of Genetic and Gradient Descent
  Learning in Recurrent Neural Networks: An Application to Speech Phoneme
  Classification.
\newblock In: Proc. of International Conference on Artificial Intelligence and
  Pattern Recognition; 2007. p. 286--293.

\bibitem{Werbos_1990}
Werbos PJ.
\newblock Backpropagation through time: what it does and how to do it.
\newblock Proceedings of the IEEE. 1990;78(10):1550--1560.

\bibitem{Hochreiter_1998}
Hochreiter S.
\newblock The vanishing gradient problem during learning recurrent neural nets
  and problem solutions.
\newblock Int J Uncertain Fuzziness Knowl-Based Syst. 1998;6(2):107--116. 

\bibitem{hochreiter1997long}
Hochreiter S, Schmidhuber J.
\newblock Long short-term memory.
\newblock Neural computation. 1997;9(8):1735--1780.

\bibitem{graves2005}
Graves A, Schmidhuber J.
\newblock Framewise phoneme classification with bidirectional LSTM and other
  neural network architectures.
\newblock Neural Networks. 2005;18:602--10. 

\bibitem{schuster1997}
Schuster M, Paliwal K.
\newblock Bidirectional recurrent neural networks.
\newblock Signal Processing, IEEE Transactions on. 1997;45:2673 -- 2681. 

\bibitem{Fan2014TTSSW}
Fan Y, Qian Y, Xie FL, Soong FK.
\newblock TTS synthesis with bidirectional LSTM based recurrent neural
  networks.
\newblock In: Fifteenth annual conference of the international speech communication association; 2014. p. 1964-- 1968.

\bibitem{graves2013hybrid}
Graves A, Jaitly N, Mohamed Ar.
\newblock Hybrid speech recognition with deep bidirectional LSTM.
\newblock In: 2013 IEEE workshop on automatic speech recognition and
  understanding. IEEE; 2013. p. 273--278.

\bibitem{vaswani2017attention}
Vaswani A, Shazeer N, Parmar N, Uszkoreit J, Jones L, Gomez AN, et~al.
\newblock Attention is all you need.
\newblock arXiv preprint arXiv:170603762. 2017.

\bibitem{wolf2020transformers}
Wolf T, Chaumond J, Debut L, Sanh V, Delangue C, Moi A, et~al.
\newblock Transformers: State-of-the-art natural language processing.
\newblock In: Proceedings of the 2020 Conference on Empirical Methods in
  Natural Language Processing: System Demonstrations; 2020. p. 38--45.

\bibitem{devlin2018bert}
Devlin J, Chang MW, Lee K, Toutanova K.
\newblock Bert: Pre-training of deep bidirectional transformers for language
  understanding.
\newblock arXiv preprint arXiv:181004805. 2018.

\bibitem{su2020application}
Su Y, Xiang H, Xie H, Yu Y, Dong S, Yang Z, et~al.
\newblock Application of BERT to Enable Gene Classification Based on Clinical Evidence.
\newblock BioMed Research International. 2020:
\newblock https://doi.org/10.1155/2020/5491963

\bibitem{dembczynski2012label}
Dembczy{\'n}ski K, Waegeman W, Cheng W, H{\"u}llermeier E.
\newblock On label dependence and loss minimization in multi-label
  classification.
\newblock Machine Learning. 2012;88(1-2):5--45.

\bibitem{zhang2018generalized}
Zhang Z, Sabuncu MR.
\newblock Generalized cross entropy loss for training deep neural networks with
  noisy labels.
\newblock arXiv preprint arXiv:180507836. 2018;.

\bibitem{hamers1989similarity}
Hamers L, et~al.
\newblock Similarity measures in scientometric research: The Jaccard index
  versus Salton's cosine formula.
\newblock Information Processing and Management. 1989;25(3):315--18.

\bibitem{furnkranz2008multilabel}
F{\"u}rnkranz J, H{\"u}llermeier E, Menc{\'\i}a EL, Brinker K.
\newblock Multilabel classification via calibrated label ranking.
\newblock Machine learning. 2008;73(2):133--153.

\bibitem{jeni2013facing}
Jeni LA, Cohn JF, De~La~Torre F.
\newblock Facing imbalanced data--recommendations for the use of performance
  metrics.
\newblock In: 2013 Humaine association conference on affective computing and
  intelligent interaction. IEEE; 2013. p. 245--251.

\bibitem{lewis1996training}
Lewis DD, Schapire RE, Callan JP, Papka R.
\newblock Training algorithms for linear text classifiers.
\newblock In: Proceedings of the 19th annual international ACM SIGIR conference
  on Research and development in information retrieval; 1996. p. 298--306.

\bibitem{lancet2020india}
Lancet T.
\newblock India under {COVID-19} lockdown.
\newblock Lancet (London, England). 2020;395(10233):1315.



\bibitem{indiaworld}
\newblock Total Coronavirus Cases in India. Worldometer. Date last  accessed: 26th, July, 2021: 
\newblock   \url{https://www.worldometers.info/coronavirus/country/india/}



\bibitem{indiapop}
Unique identification authority of {India}; 2020.
\newblock Date last  accessed: 26th, July, 2021: 
  \url{https://uidai.gov.in/images/state-wise-aadhaar-saturation.pdf}.
  
  

\bibitem{lamsal2020design}
Lamsal R.
\newblock Design and analysis of a large-scale COVID-19 tweets dataset.
\newblock Applied Intelligence. 2020; p. 1--15.

\bibitem{jack2014dynamic}
Jack RE, Garrod OG, Schyns PG.
\newblock Dynamic facial expressions of emotion transmit an evolving hierarchy
  of signals over time.
\newblock Current biology. 2014;24(2):187--192.

\bibitem{trampe2015emotions}
Trampe D, Quoidbach J, Taquet M.
\newblock Emotions in everyday life.
\newblock PloS one. 2015;10(12):e0145450.

\bibitem{cowen2017self}
Cowen AS, Keltner D.
\newblock Self-report captures 27 distinct categories of emotion bridged by
  continuous gradients.
\newblock Proceedings of the National Academy of Sciences.
  2017;114(38):E7900--E7909.

\bibitem{wintre1994developmental}
Wintre MG, Vallance DD.
\newblock A developmental sequence in the comprehension of emotions: Intensity,
  multiple emotions, and valence.
\newblock Developmental psychology. 1994;30(4):509.

\bibitem{kemper1987many}
Kemper TD.
\newblock How many emotions are there? Wedding the social and the autonomic
  components.
\newblock American journal of Sociology. 1987;93(2):263--289.

\bibitem{ekman1992argument}
Ekman P.
\newblock An argument for basic emotions.
\newblock Cognition \& emotion. 1992;6(3-4):169--200.

\bibitem{lucas2003measuring}
Lucas RE, Diener E, Larsen RJ. Measuring positive emotions; 2003.

\bibitem{friedman2014comedy}
Friedman S.
\newblock Comedy and distinction: The cultural currency of a ‘Good’sense of
  humour.
\newblock Routledge; 2014.

\bibitem{rossato2010audiences}
Rossato L, Chiaro D.
\newblock Audiences and translated humour: An empirical study.
\newblock Translation, humour and the media. 2010; p. 121--137.

\bibitem{kuipers2011politics}
Kuipers G.
\newblock The politics of humour in the public sphere: Cartoons, power and
  modernity in the first transnational humour scandal.
\newblock European Journal of Cultural Studies. 2011;14(1):63--80.

\bibitem{chey2011youmo}
Chey J.
\newblock Youmo and the Chinese sense of humour.
\newblock Humour in Chinese life and letters: Classical and traditional
  approaches. 2011; p. 1--30.

\bibitem{clasquin2001real}
Clasquin M.
\newblock Real Buddhas don't laugh: Attitudes towards humour and laughter in
  ancient India and China.
\newblock Social Identities. 2001;7(1):97--116.

\bibitem{elst2011humour}
Elst K.
\newblock Humour in Hinduism.
\newblock Humour and Religion: Challenges and Ambiguities London: Bloomsbury
  Academic. 2011; p. 35--53.

\bibitem{chandra2019langevin}
Chandra R, Jain K, Deo RV, Cripps S.
\newblock Langevin-gradient parallel tempering for Bayesian neural learning.
\newblock Neurocomputing. 2019;359:315--326.

\bibitem{chandra2021revisiting}
Chandra R, Jain M, Maharana M, Krivitsky PN.
\newblock Revisiting {Bayesian} Autoencoders with {MCMC}.
\newblock arXiv preprint arXiv:210405915. 2021.

\bibitem{chandra2021bayesian}
Chandra R, Bhagat A, Maharana M, Krivitsky PN.
\newblock Bayesian graph convolutional neural networks via tempered {MCMC}.
\newblock arXiv preprint arXiv:210408438. 2021.



\end{thebibliography}







\end{document}